%% file: ARXIV_VER.tex
\documentclass[a4paper]{article}
\usepackage[margin=35mm]{geometry}
\usepackage{amssymb}
\usepackage{url}
\usepackage{arydshln}
\usepackage[normalem]{ulem}
\usepackage{amsmath}
\usepackage{breakcites}
\usepackage{graphicx,xcolor}
\usepackage{placeins}

\usepackage{subcaption}
\usepackage{booktabs}

\usepackage{tikz}
\usetikzlibrary{positioning, shapes, shadows, arrows}
\usetikzlibrary{decorations.pathreplacing}
\tikzstyle{abstract}=[circle, draw=black, fill=white]
\tikzstyle{labelnode}=[circle, draw=white,opacity=.2,text opacity=1]
\tikzstyle{invisiblenode}=[circle,dashed, inner sep=1pt,circle split,line width=1mm,minimum size=1.5cm]
\tikzstyle{line} = [draw, -latex']

\let\vec\mathbf
\definecolor{colrev}{rgb}{1.0, 0.0, 1.0}   

\newcommand{\citep }[1]{\cite{#1}}

\renewcommand{\textcolor}[2]{#2}

\newcommand\blfootnote[1]{%
  \begingroup
  \renewcommand\thefootnote{}\footnote{#1}%
  \addtocounter{footnote}{-1}%
  \endgroup
}

\usepackage[T1]{fontenc} 
\usepackage[utf8]{inputenc}

\providecommand{\keywords}[1]
{
  \small	
  \textbf{\textit{Keywords---}} #1
}

\begin{document}

\title{A survey on modern trainable activation functions}
\date{}

\author{Andrea Apicella$^1$, Francesco Donnarumma$^2$, Francesco Isgr\`o$^1$ \\
and Roberto Prevete$^1$\\
\small $^1$Dipartimento di Ingegneria Elettrica e delle Tecnologie dell'Informazione\\\small Universit\`a di Napoli Federico II, Italy\\
\small$^2$Istituto di Scienze e Tecnologie della Cognizione\\\small CNR, Italy
}
\maketitle
\begin{abstract}
In neural networks literature, there is a strong interest in identifying and defining activation functions which can improve neural network performance. 
In recent years  there has been a renovated interest of the scientific community in investigating activation functions which can be trained during the learning process, usually referred to as \textit{trainable}, \textit{learnable}  or \textit{adaptable} activation functions. 
They appear to lead to better network performance. Diverse and heterogeneous models of trainable activation function have been proposed in the literature. 
In this paper, we present a survey of these models. Starting from a discussion on the use of the term ``activation function'' in literature, we propose a taxonomy of trainable activation functions, highlight common and distinctive proprieties of recent and past models, and discuss main advantages and limitations of this type of approach. 
We show that many of the proposed approaches are equivalent to adding neuron layers which use fixed (\textbf{non-trainable}) activation functions and some simple local rule that constraints the corresponding weight layers. 
\end{abstract}


\thispagestyle{empty}
\blfootnote{This is a preprint of the survey published in \textit{Neural Networks}, Volume 138, June 2021, Pages 14-32\\Please cite the published version \url{https://doi.org/10.1016/j.neunet.2021.01.026}}

\input{introduction.tex}

\input{taxonomy.tex}

\input{fixed_shape.tex}
\input{learn_shape.tex}
\input{discussion.tex}

\input{conclusions.tex}
\input{TAB/results.tex}

\bibliographystyle{apalike} 
\bibliography{bibliografia}

\end{document}

%% file: introduction.tex
\section{Introduction}
\label{sec:introduction}
\begin{figure}
\centering
\includegraphics[width=0.8\textwidth]{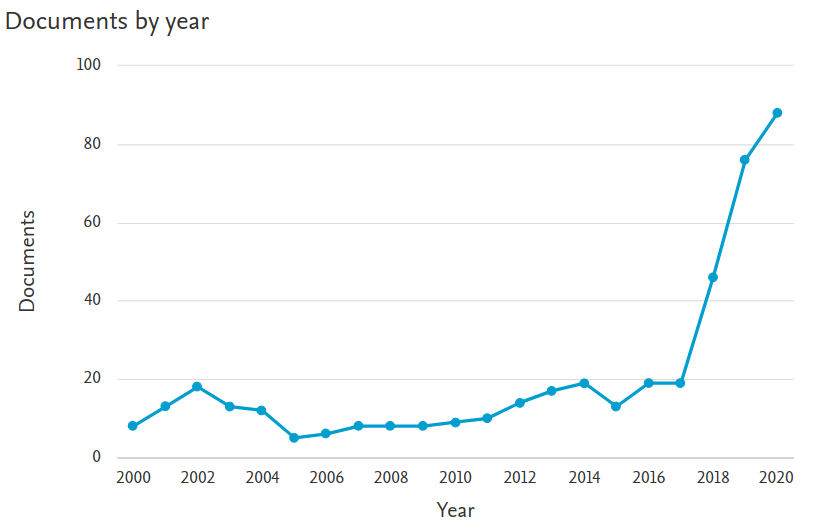}
\caption{Number of papers by year on trainable activation functions. Source: Scopus, query: ((((("trainable activation  function"  OR  "learnable activation function")  OR  "adaptive activation function")  OR  "adaptable activation function")  AND  "neural networks"))}
\label{fig:documents}
\end{figure}

The introduction of new activation functions has contributed to renewing the interest of the scientific community \textcolor{black}{in neural networks,} having a central role for the expressiveness of artificial neural networks.  
For example, the use of ReLU \citep{glorot2011}, Leaky ReLU \citep{maas2013}, parametric ReLU \citep{he2015}, and similar activation functions (for example, \citep{dugas2000,clevert2015}), has shown to improve the network performances significantly, \textcolor{black}{thanks to properties such as the no saturation feature} that helps to avoid typical learning problems as vanishing gradient \citep{bishop2006}. 
Thus, individuating new activation functions that \textcolor{black}{can potentially improve} the results is still an open field of research.
In this research line, an investigated and promising approach is the possibility to determine an \emph{appropriate}  activation function by learning. 
In other words, the key idea is to involve the activation functions in the learning process together (or separately) with the other parameters of the network such as weights and biases, thus obtaining a \textit{trained activation function}. 
\textcolor{black}{In literature} we usually find the expression  ``trainable activation functions'', however  the expressions ``learneable'', ``adaptive'' or ``adaptable'' activation functions are also used, see, for example, \citep{scardapane2018,apicella2019,qian2018}. 
Many and heterogeneous trainable activation function models have been proposed \textcolor{black}{in literature}, and in recent years \textcolor{black}{there has been} a particular interest in this topic, see Figure \ref{fig:documents}. 

In this paper, we present a survey of trainable activation functions in the neural network domain, highlighting general and peculiar proprieties of recent and past approaches.
In particular, we examine trainable activation functions in the context of feed-forward neural networks, although many of the approaches that we are going to discuss can also be applied to recurrent neural networks. 
In the first place, the relevant and critical properties of these approaches are isolated. Taking into consideration this analysis, we propose a taxonomy that characterizes and classifies these functions according to their definition. 
Moreover, we show that many of the proposed approaches are equivalent to adding neuron layers which use fixed activation functions (\textbf{non-trainable} activation functions) and some simple \textcolor{black}{local rule that constrains} the corresponding weight layers. 
Also, based on this taxonomy, we discuss the expressivity of these functions and the performances that were achieved when neural networks with trainable activation functions were used.

In order to explain and \textcolor{black}{better analyze} the various models of trainable activation function, we start describing what is usually meant by the expressions ``feed-forward neural network'' and ``activation function'', summarizing the main non-trainable (fixed) activation functions proposed \textcolor{black}{in literature} so far.

\subsection*{Definitions and symbols}
Multi-Layer Feed Forward (MLFF) networks are composed of $N$ elementary computing units (neurons), which are organized in $L>1$ layers.
The first layer of an MLFF network is composed of $d$ input variables. 
Each neuron $i$ belonging to a layer $l$, with $1 \leq l \leq L$, may receive connections from all the neurons (or input variables in case of $l=1$) of the previous layer $l-1$. 
Each connection is associated with a real value called \textit{weight}. 

The flow of computation proceeds from the first layer to the last layer (\textit{forward propagation}).  
The last neuron layer is called \textit{output} layer, the remaining neuron layers are called \textit{hidden} layers. 
The computation of a neuron $i$ belonging to the layer $l$ corresponds to a two-step process: \textcolor{black}{the first step is the computation of the the neuron input $a_{i}^l$ is computed; the second step is the computation of the neuron output $z_i^{l}$.} 
The neuron input $a_{i}^l$ is usually constructed as a linear combination of its incoming input values, corresponding to the output of the previous layer: $a_{i}^{l}=\sum_j w_{ij}^l z^{l-1}_j + b^l_{i}$ where $w_{ij}^l$ is the weight of the connection going to the neuron $j$ belonging to the layer $l-1$ to the neuron $i$ belonging to the layer $l$, $b^l_{i}$ is a parameter said $bias$, $z^{l-1}_j$ is the output of the neuron $j$ belonging to the layer $l-1$ (or the input variables, if $l=1$), and $j$ runs on the indexes of the neurons of the layer $l-1$ (or the input variables, if $l=1$) which send connections to the neuron $i$. 
In a standard matrix notation, $a_{i}^{l}$ can be expressed as $a_{i}^{l}= \vec{w}^{lT}_i \vec{z}^{l-1}+b_i$, where $\vec{w}^l_i$ is the weight column vector \textcolor{black}{associated with} the neuron $i$ belonging to the layer $l$ and $\vec{z}^{l-1}$  is the column vector corresponding to the output of the neurons belonging to the layer $l-1$. If $l=1$ the vector $\vec{z}^{l-1}$ corresponds to the input variables. 
The neuron output 
$z_{i}^l$ is usually computed by a differentiable, non linear function $f(\cdot)$: $z_{i}^l=f(a_{i}^l)$.
In this network model, the nonlinear functions $f(\cdot)$ are generally chosen as simple \textit{fixed} functions such as the \textit{logistic sigmoid} or the \textit{tanh} functions, and they are usually called \textit{activation functions}.

\subsection*{Activation functions: a brief historical excursus}
The expression \textit{activation function} \textcolor{black}{has not always been used} with today's meaning, since other expressions have been used in literature, as \textit{transfer function} or \textit{output function}. 
In some cases, \textit{transfer function} is used as synonym of  \textit{activation function} (such as in \citep{hagan1996}) but, in other cases, there is a clear distinction between the different forms. For instance, in a well known survey \citep{duch1999}, the two forms ``activation function'' and ``output function'', assume different meanings; more precisely: ``the activation function determines the total signal a neuron receives. The value of the activation function is usually scalar and the arguments are vectors. The second function determining neuron's signal processing is the output function [...], operating on scalar activations and returning scalar values. 
Typically a squashing function is used to keep the output values within specified bounds. 
These two functions together determine the values of the neuron outgoing signals. 
The composition of the activation and the output function is called the transfer function''. 
In a nutshell, \citep{duch1999} distinguish among:
\begin{itemize}
\item \textit{activation function} $I(\vec{z})$: an internal transformation of the input values $\vec{z}$. The most common artificial neuron model makes a weighted sum of the input values, that is $I(\vec{z})= \vec{w}^T \vec{z}+b$ where $\vec{w},b$ are said neuron parameters (or weights) and the bias, respectively. However, in the recent literature the statement \textit{activation function} loses this meaning, as described in the following;
\item \textit{output function} $o(a)$: a function which returns the output value of the neuron using the activation value $a=I(\vec{z})$, i.e. $o : a\in \mathbb{R} \to o(a)\in \mathbb{R}$. However, the largest part \textcolor{black}{of recent literature} usually refers to this function as  \textit{activation function}, so giving a different meaning respect to the word \textit{activation};
\item \textit{transfer function} $T(\vec{z})$: the composition of output function and activation function, that is $T(\vec{z})=o(I(\vec{z}))$.
\end{itemize}
This distinction between activation, output and transfer function was not used in previous research works, such as   
in \citep{haykin1994}, where an activation function is  ``[...] a squashing function in that it squashes (limits) the permissible amplitude range of the output signal to some finite value''. 
Here, it is clear that the ``activation function'' is what \citep{duch1999} \textcolor{black}{define as} ``output function''. 
The terminology and the formalization used in \citep{haykin1994} \textcolor{black}{are} the most used \textcolor{black}{in literature.} 
For example,  \citep{dasgupta1993} defines an activation function as a member of ``a  class of real-valued functions, where each function is defined on some subset of $\mathbb{R}$''. 

In \citep{goodfellow2016} an activation function is ``a fixed nonlinear function''. 
The nonlinearity requirement comes from \citep{cybenko1989,hornik1989} where it is shown that  the activation functions have to be non-constant, bounded and monotonically-increasing continuous to ensure the network’s universal approximator property (see Section \ref{sec:fixsh}). 
In \citep{muller2012} the activation function is introduced as ``A transfer function $f_i$ [...] defined for each [network] node $i$, which determines the state of the node as a function composed of its bias, the weights of incoming links, and the states of nodes connected to it'', so using again the term \textit{transfer} and \textit{activation} in an interchangeable manner. 
In \citep{eldan2016} an activation function is clearly defined as any $f: \mathbb{R} \to \mathbb{R}$ function. 
All these definitions agree in defining an activation function as a functional mapping between two subsets of the real \textcolor{black}{numbers, provided} that this function meets  suitable requirements to guarantee the MLFF network’s universal approximator property.

An exception seems to be \citep{hagan1996}, where  the expressions ``activation function'' and ``transfer function'' are used indistinctly as synonyms of what the authors in \citep{duch1999} call ``output function''. 
More precisely,  in \citep{hagan1996} the authors define the activation/transfer function as the function which ``produces the scalar neuron output [...]. The transfer function [...] may be a linear or a nonlinear function [...]. A particular transfer function is chosen to satisfy some specification of the problem that the neuron is attempting to solve''.

In another direction, part \textcolor{black}{of literature} about trainable activation functions loses the concept of activation function as reported in works such as \citep{dasgupta1993,haykin1994,goodfellow2016}, proposing instead new neuron architectures which seem to work without a clearly-defined activation function, generating an output that is different from a simple non-linearity applied to a linear combination of the neuron inputs. 
These approaches seem to change all the internal neuron behavior, so, instead of trainable activation functions, if we want to keep the distinction reported in \citep{duch1999}, we should refer to these as \textcolor{black}{trainable transfer functions.} 
\bigskip

\textcolor{red}{Summarising, the behaviour of an artificial neural network (ANN) is typically characterized by the elementary computations of each neuron \citep{leshno1993,bishop1995neural,bishop2006,njikam2016novel}, and the elementary computational process of each neuron is generally a two-step process \citep{bishop1995neural,bishop2006}. In the first one, a real value is computed from the neuron input values and the incoming connection weights. Usually, this value is computed as a weighted linear combination of the neuron’s input, the standard inner-product in $\mathbb{R}^d$ between the vectors $\vec{x}$ and $\vec{w}$.
Different definitions have been proposed in the literature (see, for example, \citep{chen1991orthogonal}). In the second step, a scalar output value is computed by a Real activation function, $f : \mathbb{R} \longrightarrow \mathbb{R}$. 
From a historical point of view, different definitions of activation functions can be found in the literature as we discussed previously in this Section.  However, the current literature still identifies two distinct steps \citep{bishop1995neural,bishop2006,scardapane2018} in the computation at the neuron level. Therefore, in this paper, in order to highlight this two-step process, unless otherwise specified, we distinguish among 1) an input function $I : \mathbb{R}^d \longrightarrow \mathbb{R}$ characterising the first step of the neuron computation, 2) an activation function $f: \mathbb{R}\longrightarrow \mathbb{R}$ which characterises the second step, and 3) a transfer function $T: \mathbb{R}^d \longrightarrow \mathbb{R}$ as the composition of $I$ and $f$, $T=f(I(z))$, characterising the whole neuron computation. In this way, a neuron output depends directly on the incoming input values (input values of the whole network or outputs of other neurons) and connection weights only.}
\textcolor{red}{
Furthermore, in the context of non-fixed activation functions the words \textit{trainable}, \textit{learnable} and \textit{adaptable} are used as synonyms.}

In this work, we give three main contributions; the first one is a survey on the current state of the art of trainable activation functions and the obtained results. The second one consists of highlighting relevant and critical properties of these approaches. Taking into consideration this analysis, we propose a taxonomy that characterizes and classifies these functions according to their definition.
The last contribution is to show that, in many cases, using a trainable activation is equivalent to using a deeper neural network model with additional constraints on the parameters.

The work is so organized as follows: in Section \ref{sec:aft} we propose a possible \textcolor{black}{activation function taxonomy}, distinguishing them between fixed and trainable ones. In Section \ref{sec:fixsh} we give a brief summary of the most used fixed activation functions, while in Section~\ref{sec:trsh} and \ref{sec:trainableTransferFunctions} we make a survey of the state of art of the trainable activation functions; in Section \ref{sec:disc} we discuss the obtained performances in literature. Section~\ref{sec:conclusions} is left to final remarks.



%% file: taxonomy.tex
\section{A taxonomy of activation functions}
\label{sec:aft}
In this work, we propose a possible taxonomy of the main activation functions presented \textcolor{black}{in literature,} see Figure \ref{fig:taxonomy}. 
As stated in the previous section, we focus on activation functions as defined in \citep{haykin1994,goodfellow2016}, i.e., the output of a neuron is computed by a two-step process: first the input of the neuron is computed by a functional mapping from $\mathbb{R}^d$ to $\mathbb{R}$ (usually a weighted sum), then the output (or activation) of the neuron is computed by the activation function which is a functional mapping from $\mathbb{R}$ to $\mathbb{R}$. 
This way to compute the neuron output is widely used in literature and de facto it is the standard in artificial neural networks. However, a number of significant neural network models which implement different approaches have been presented in literature.
Among these models we have isolated a subset of them which can be interpreted as ``trainable non standard neuron  definitions''. 
In our taxonomy we put them  as a distinct class, and we \textcolor{black}{discuss trainable non standard neuron  definitions} in Section \ref{sec:trainableTransferFunctions}.

\input{FIGURE/tasson.tex} 

The primary classification is based on the possibility of modifying the activation function shape during the training phase. So, one can isolate two main categories:
\begin{itemize}
    \item \textbf{fixed-shape activation functions}: all the activation functions with a fixed shape, for example, all the classic activation functions used in neural network literature, such as sigmoid, tanh, ReLU, \textcolor{black}{fall into this category.}
    However, since the introduction of rectified functions (as ReLU) can be considered a turning point \textcolor{black}{in literature} contributing to improving the neural network performances significantly and increasing the interest of the scientific community, we can further divide this class of \textcolor{black}{functions into:}
    \begin{itemize}
        \item rectified-based function: all the functions belonging to the rectifier family, such as ReLU, LReLU, etc.
        \item classic activation function: all the functions that are not in the rectifier family, such as the sigmoid, tanh, step functions.
    \end{itemize}
    
    \item \textbf{trainable activation functions}: this class contains all the activation functions the shape of which is learned during the training phase.  
    The idea behind this kind of functions is to search a good function shape using knowledge given by the training data. 
    However, we will show that several trainable activation functions can be reduced to classical feed-forward neural subnetworks composed of neurons equipped with classical fixed activation functions, grafted into the main neural network only by adding further layers. 
    In other terms, a neural network architecture equipped with trainable activation functions can have a similar (if not the same) behaviour of a deeper network architecture equipped with just classic (fixed) activation functions, in some cases by adding simple constraints on the network parameters, e.g., by fixing some weights or sharing them (as in convolutional networks) and by proper arranging of the layers.
    
    Taking into account all these considerations, \textcolor{black}{among all} the trainable activation functions described into the literature we can isolate two different families:
        \begin{itemize}
            \item \textit{parameterized standard functions}: in this case, we consider all the trainable functions derived from standard fixed activation functions with the addition of a set of trainable parameters. 
            In other words, they are defined as a parameterized version of a standard fixed function whose parameter values are learned from data. We will see later that several of these functions can be expressed in terms of subnetworks. 
            \item \textit{Functions based on ensemble methods}:  they are defined by mixing several distinct functions. A common way to mix different functions is combining them \textcolor{black}{linearly, i.e.,} the final activation functions are modelled in terms of linear combinations of one-variable functions. We group together all these activation functions in a subclass that we named \textit{linear combination of one-to-one functions}. In this case, these one-variable functions can, in turn, have additional parameters. 
            Many of these approaches can be expressed in terms of MLFF subnetworks which receive just one single input value, as we will discuss in Section \ref{sec:LinearCombinationOneToOneFunctions}. By contrast, some activation functions are already proposed, in the original papers, in terms of one or more sub-networks which can be in turn described as a linear combination of one-to-one functions. 
            Moreover, other functions are modelled in an analytic form. We will show that also in this case, many of these functions can be modelled as subnetworks nested into the main network architecture.
        \end{itemize}
    \end{itemize}
    
The next two sections describe the distinctive features of activation function classes which have been isolated by our taxonomy and discuss some advantages and limitations of the different approaches.  

%% file: FIGURE/tasson.tex
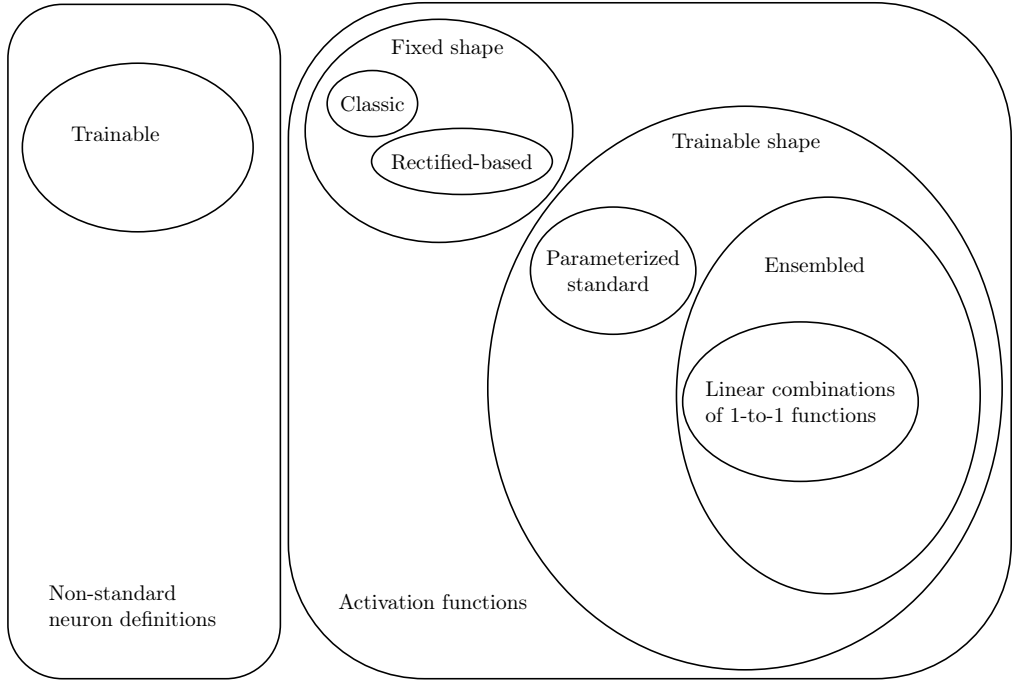
\begin{figure}[t]
\centering

\scalebox{.8}{
\tikzset{every picture/.style={line width=0.75pt}} 

\begin{tikzpicture}[x=0.75pt,y=0.75pt,yscale=-1,xscale=1]

\draw  [color={rgb, 255:red, 0; green, 0; blue, 0 }  ,draw opacity=1 ] (176,86) .. controls (176,39.06) and (214.06,1) .. (261,1) -- (542,1) .. controls (588.94,1) and (627,39.06) .. (627,86) -- (627,341) .. controls (627,387.94) and (588.94,426) .. (542,426) -- (261,426) .. controls (214.06,426) and (176,387.94) .. (176,341) -- cycle ;
\draw   (186.48,81.53) .. controls (186.48,42.79) and (223.85,11.39) .. (269.95,11.39) .. controls (316.04,11.39) and (353.41,42.79) .. (353.41,81.53) .. controls (353.41,120.27) and (316.04,151.67) .. (269.95,151.67) .. controls (223.85,151.67) and (186.48,120.27) .. (186.48,81.53) -- cycle ;
\draw   (200.19,64.39) .. controls (200.19,52.91) and (212.83,43.6) .. (228.42,43.6) .. controls (244,43.6) and (256.64,52.91) .. (256.64,64.39) .. controls (256.64,75.86) and (244,85.17) .. (228.42,85.17) .. controls (212.83,85.17) and (200.19,75.86) .. (200.19,64.39) -- cycle ;

\draw   (227.89,100.76) .. controls (227.89,89.28) and (253.15,79.97) .. (284.3,79.97) .. controls (315.45,79.97) and (340.71,89.28) .. (340.71,100.76) .. controls (340.71,112.23) and (315.45,121.54) .. (284.3,121.54) .. controls (253.15,121.54) and (227.89,112.23) .. (227.89,100.76) -- cycle ;

\draw   (1,36) .. controls (1,17.22) and (16.22,2) .. (35,2) -- (137,2) .. controls (155.78,2) and (171,17.22) .. (171,36) -- (171,392) .. controls (171,410.78) and (155.78,426) .. (137,426) -- (35,426) .. controls (16.22,426) and (1,410.78) .. (1,392) -- cycle ;
\draw   (10,92) .. controls (10,62.73) and (42.24,39) .. (82,39) .. controls (121.76,39) and (154,62.73) .. (154,92) .. controls (154,121.27) and (121.76,145) .. (82,145) .. controls (42.24,145) and (10,121.27) .. (10,92) -- cycle ;
\draw   (300.41,243.23) .. controls (300.41,145.38) and (372.26,66.06) .. (460.88,66.06) .. controls (549.51,66.06) and (621.36,145.38) .. (621.36,243.23) .. controls (621.36,341.07) and (549.51,420.4) .. (460.88,420.4) .. controls (372.26,420.4) and (300.41,341.07) .. (300.41,243.23) -- cycle ;
\draw  [color={rgb, 255:red, 0; green, 0; blue, 0 }  ,draw opacity=1 ] (327.02,169.45) .. controls (327.02,147.35) and (350.13,129.44) .. (378.63,129.44) .. controls (407.13,129.44) and (430.24,147.35) .. (430.24,169.45) .. controls (430.24,191.54) and (407.13,209.45) .. (378.63,209.45) .. controls (350.13,209.45) and (327.02,191.54) .. (327.02,169.45) -- cycle ;
\draw  [color={rgb, 255:red, 0; green, 0; blue, 0 }  ,draw opacity=1 ] (418.14,247.9) .. controls (418.14,179.04) and (460.56,123.21) .. (512.89,123.21) .. controls (565.22,123.21) and (607.65,179.04) .. (607.65,247.9) .. controls (607.65,316.77) and (565.22,372.6) .. (512.89,372.6) .. controls (460.56,372.6) and (418.14,316.77) .. (418.14,247.9) -- cycle ;
\draw  [color={rgb, 255:red, 0; green, 0; blue, 0 }  ,draw opacity=1 ] (422,251.8) .. controls (422,224.11) and (454.91,201.66) .. (495.5,201.66) .. controls (536.09,201.66) and (569,224.11) .. (569,251.8) .. controls (569,279.49) and (536.09,301.94) .. (495.5,301.94) .. controls (454.91,301.94) and (422,279.49) .. (422,251.8) -- cycle ;

\draw (284.3,100.76) node   [align=left] {Rectified-based};
\draw (228.42,64.39) node   [align=left] {Classic};
\draw (275.19,30.1) node   [align=left] {Fixed shape};
\draw (266.22,377.5) node   [align=left] {Activation functions};
\draw (25,366) node [anchor=north west][inner sep=0.75pt]   [align=left] {Non-standard\\neuron definitions};
\draw (39,77) node [anchor=north west][inner sep=0.75pt]   [align=left] {Trainable};
\draw (461.69,88.92) node   [align=left] {Trainable shape};
\draw (378.63,169.45) node   [align=left] {Parameterized\\ \ \ \ standard};
\draw (504.43,165.81) node   [align=left] {Ensembled};
\draw (495.5,251.8) node   [align=left] {Linear combinations\\of 1-to-1 functions};

\end{tikzpicture}
} 

\caption{A proposed taxonomy of the activation functions proposed \textcolor{black}{in Neural Networks literature.}}
\label{fig:taxonomy}
\end{figure}

%% file: fixed_shape.tex
\section{Fixed-shape activation functions}
\label{sec:fixsh}
With the expression \textcolor{black}{``fixed-shape activation functions'',} we indicate all the activation functions which are defined without parameters that can be modified during the training phase.

Since many trainable activation functions \textcolor{black}{in literature} are proposed as a combination or variation of fixed-activation functions, this section presents a brief description of the main fixed-shape activation functions used in neural networks. Several studies which compare different fixed activation functions have been made over the years, see for example  \citep{sibi2013,xu2015,xu2016,pedamonti2018,nwankpa2018}.
 
This section is divided into two parts. The former is dedicated to the classic activation functions used primarily in the past. The latter is about ReLU and its possible improvements given by changing its basic shape.

\subsection{Classic activation functions}
\label{sec:act_fun}
\begin{figure}
    \centering
    \includegraphics[width=\textwidth]{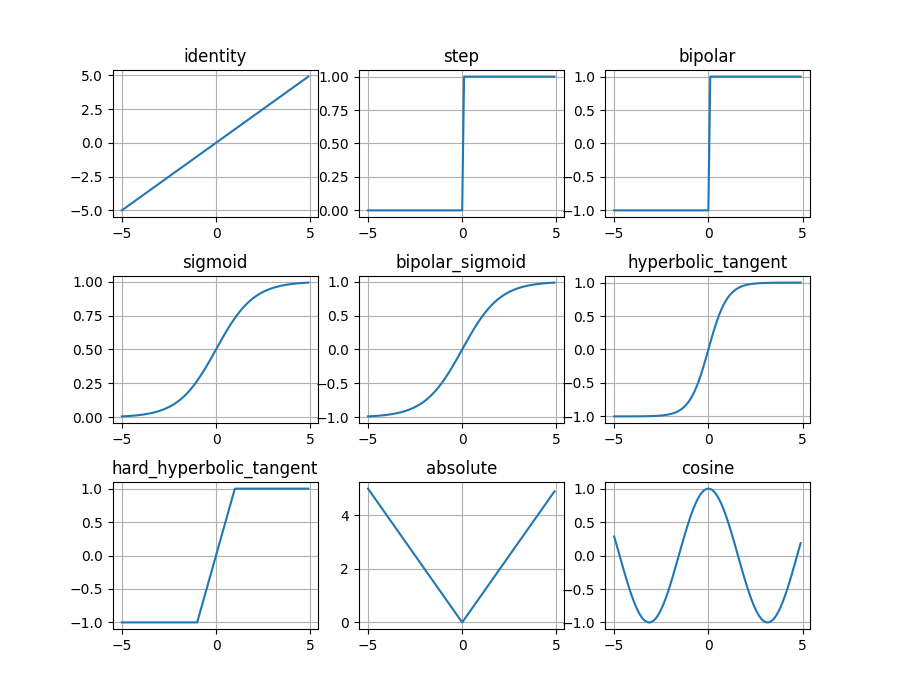}
    \caption{\textcolor{red}{Examples of classic activaton functions}}
    \label{fig:classic_af}
\end{figure}

\input{TAB/fixactfun.tex}

A short and partial list of the most common activation functions used in Neural Network literature is given in Table \ref{tab:act_fun} \textcolor{red}{and shown in Figure \ref{fig:classic_af}}. 

In \citep{cybenko1989,hornik1989} it is shown that any continuous function defined on a compact subset, can be approximated arbitrarily well by a feed-forward network with a single hidden layer (\textit{shallow network}), provided that the number of hidden neurons is sufficiently large and the activation functions are non-constant, bounded and monotonically-increasing continuous functions. 
This theorem demonstrated that activation functions like the identity function or any other linear function, used for example in early Neural Networks as ADALINE or MADALINE \citep{widrow1960,widrow1990}, cannot approximate any continuous function.
Therefore, for many years, bounded activation functions such as sigmoid \citep{cybenko1988} or hyperbolic tangent \citep{chen1990}  have been the most used activation functions for neural networks.

Over the years, several studies showed how bounded activation functions could reach excellent results, especially in shallow network architectures (see for example \citep{glorot2011,pedamonti2018}).
Unfortunately, the training of networks equipped with these functions suffers from the \textit{vanishing gradient problem} (see \citep{bengio1994}) when networks with many layers (Deep Neural Networks, DNN) are used, compromising the network training. 
In \citep{pinkus1999,sonoda2017} it was shown that the requirements specified in \citep{cybenko1989,hornik1989} for the activation functions to give to a network the universal approximation property were too strong, showing that also neural networks equipped with unbounded but non-polynomial activation functions (e.g. ReLU, \citep{nair2010}) are universal approximators. 
Furthermore, unbounded activation functions seem to attenuate the vanishing gradient problem (see \citep{nair2010}), opening new frontiers in neural networks and machine learning research.

\subsection{Rectifier-based activation functions}
\begin{figure}
    \centering
    \includegraphics[width=\textwidth]{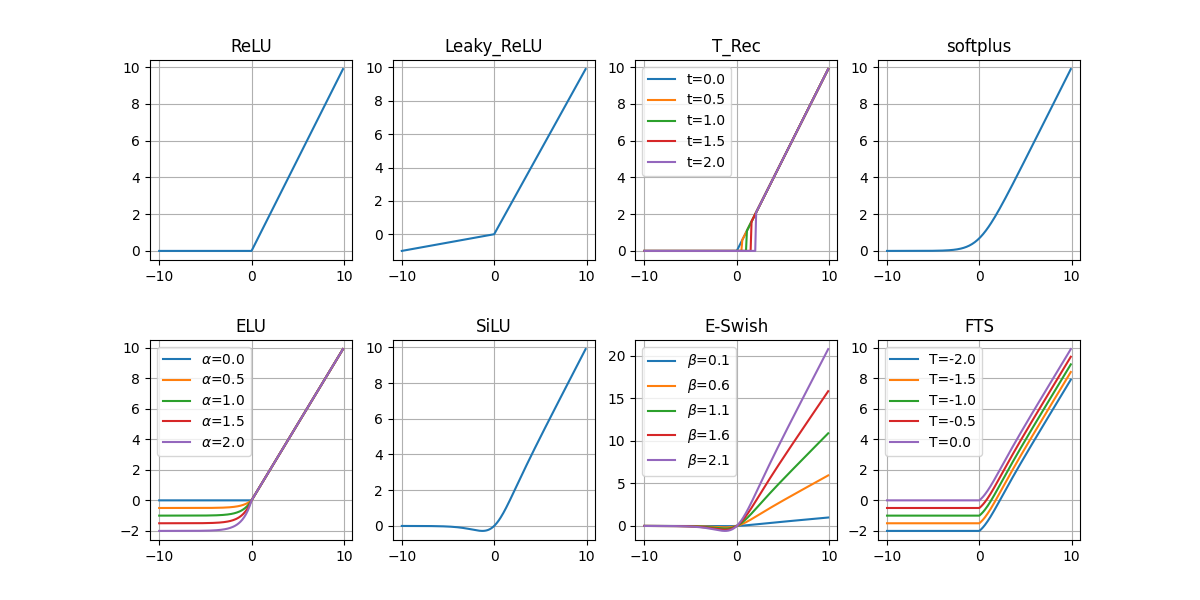}
    \caption{\textcolor{red}{Examples of rectifier-based activaton functions}}
    \label{fig:rectifier_af}
\end{figure}
Over the last years, many different activation functions have been proposed, most of which inspired by the success obtained by ReLU \citep{glorot2011}, and therefore based on a similar shape, with small variations, compared to the original function. 
This kind of activation functions is the standard \textit{de facto} in current neural network architectures, overcoming other such classic functions as sigmoid and tanh used in the past. 
One of the first studies that showed the performance improvements of networks equipped with rectifier-based activation functions was \citep{glorot2011}, where DNNs equipped with ReLU activation functions improved the performances if compared to networks with sigmoid units.
As stated above, the main benefit of using rectified activation functions is to avoid the vanishing gradient problem \citep{bengio1994}, which has been one of the main problems for deep networks for many years. Various efforts continue to be made in the scientific community to find new activation functions to improve neural network performances. 
However, the research on classic activation functions has not come to an end, for example, properties on these functions are still discussed in recent studies \citep{gulcehre2016,xu2016}.  

In the remainder of this subsection, we identify the main characteristics of ReLU and ReLU-family functions, \textcolor{red}{some of which are shown in Figure \ref{fig:rectifier_af}}.

\paragraph{\textbf{ReLU}}
This function, defined as $\text{ReLU}(a)=\max(0,a)$, has significant positive features:
\begin{itemize}
    \item It alleviates the vanishing gradient problem being not bounded in at least one direction;
    \item  It facilitates \textit{sparse coding} \citep{tessitore2011designing,montalto2015linear}, as the percentage of neurons that are really active at the same time is usually very low.
    The benefits of sparsity are described in \citep{glorot2011} and can be resumed in a better dimensionality of the representation and a more significant invariance to slight input changes.
\end{itemize}

However, ReLU function is not clear from defects, \textcolor{black}{such as} the ``dying'' ReLU problem \citep{maas2013} or the non-differentiability at zero, that appears when a large negative bias is learned causing the output of the neuron to be always zero regardless of the input. 
Consequently, in the following, we discuss some ReLU variants.

\paragraph{\textbf{Leaky ReLU}}:
One of the earliest rectified-based activation function based on ReLU was LReLU \citep{maas2013}. 

LReLU function was an attempt to alleviate the potential problems of the ReLU mentioned above. It is defined as:
\begin{equation*}
\text{LReLU}(a) = \begin{cases}
 a & \text{ if }a > 0\\
 0.01 \cdot a & \text{otherwise.}
\end{cases}
\end{equation*}
A Leaky Rectifier Activation Function allows the unit to give a small gradient when the unit is saturated and not active, i.e., when $a \leq 0$. 
However, the authors show empirically that Leaky rectifiers perform nearly identically to standard rectifiers, resulting in a negligible impact on network performances.

A randomized version (Randomized Leaky ReLU), where the weight value for $a$ is sampled by a uniform distribution $U(l,u)$ with $0\leq l<u<1$ was also proposed in \citep{xu2015}.

\paragraph{\textbf{Truncated Rectified}}
Authors in \citep{konda2014}, tackle the problem to find alternatives to ReLU focusing on a particular type of DNN (Autoencoders), starting from the observation that this type of network tends to have large negative bias, which can have several side effects on the learning process (see \citep{konda2014} for further details).
From this observation, the authors propose the Truncated Rectified as activation function which can be defined as:
\begin{equation*}
\text{TRec}_{t}(a)= \begin{cases}
 a & \text{ if }a > t\\
 0 & \text{otherwise.}
\end{cases}
\end{equation*}

Note that the $t$ point is a non-continuity point, unlike ReLU in which the threshold point (which is $0$)  is only a non-differentiable point. Authors use TRec only during training, and then replace it with ReLU during testing. The authors make this choice in order to obtain both sparse coding and the minimizing of the error function without any kind of weight regularization.

\paragraph{\textbf{Softplus}}
Introduced by \citep{dugas2000}, the $softplus$ function can be seen as a smooth approximation of ReLU function. 
It is defined as
\begin{equation*}
    \text{softplus}(a) = \log{(1+\exp (a))}
\end{equation*}
The smoother form and the lack of points of non-differentiation could suggest a better behavior and an easier training as an activation function. However, experimental results \citep{glorot2011} tend to contradict this hypothesis, suggesting that ReLU properties can help supervised training better than softplus functions.

\paragraph{\textbf{Exponential Linear Unit (ELU)}}
Introduced in \citep{clevert2015}, the ELU is an activation function that keeps the identity for positive arguments but with non-zero values for negative ones.
It is defined as:
 \begin{equation*}
\text{ELU}(a)=  \begin{cases}
 a & \text{ if }a > 0\\
 \alpha \cdot (\exp (a) -1) & \text{otherwise.}
\end{cases}
\end{equation*}
where $\alpha$ controls the value for negative inputs. 
The values given by ELU units push the mean of the activations closer to zero, allowing a faster learning phase (as showed in \citep{clevert2015}), at the cost of an extra hyper-parameter ($\alpha$) which requires to be set.

\paragraph{\textbf{Sigmoid-weighted Linear Unit}}
Originally proposed in \citep{elfwing2018}, Sigmoid-weighted Linear Unit is a sigmoid function weighted by its input, i.e.:
\begin{equation*}
\text{SiLU}(a) = a \cdot \text{sig}( a)
\end{equation*}
In the same study, the derivative of SiLU is also proposed as activation function, i.e.:
\begin{equation*}
\text{dSiLU}(a) = \text{sig}(a)\big(1+ a(1-  \text{sig}( a))\big)
\end{equation*}
These functions have been tested on reinforcement learning tasks.
Moreover, further applications are given in \citep{ramachandran2017} where the same function is tested with the name of \text{Swish}-1.

\paragraph{\textbf{E-swish}}
A SiLU variation is proposed in \citep{alcaide2018} by adding a multiplicative coefficient to the SiLU function, and obtaining:
\begin{equation*}
\text{E-swish}_\beta (a) = \beta \cdot a \cdot \text{sig}(a)
\end{equation*}
with $\beta \in \mathbb{R}$. 
The function name comes from the Swish activation function, a trainable version of SiLU function proposed in \citep{ramachandran2017} (see Section \ref{sec:swish} for further details). 
However, E-Swish has no trainable parameters, leaving to the user the tuning of the $\beta$ parameter. 
The authors consider the $\beta$ parameter \textcolor{black}{as a hyper-parameter} to be found by a search procedure. 

\paragraph{\textbf{Flatten-T Swish}}
The function described in \citep{chieng2018} has properties of both ReLU and sigmoid, combining them in a manner similar to the Swish function.
\begin{equation*}
    \text{FTS}(a)= \begin{cases}
    a \cdot \frac{1}{1+ \exp(-a)} + T, & \text{ if } x \geq 0\\
    T, & \text{otherwise.}
    \end{cases}
\end{equation*}
When $T=0$ the function becomes $\text{ReLU}(a) \cdot \text{sig}(a)$, a function similar to \text{Swish}-1, where the ReLU function substitutes the identity. $T$ is an additional fixed threshold value to allow the function to return a negative value (if $T<0$), differently from the classic ReLU function. The authors plan to propose a method to learn the parameter $T$ in future work. 

%% file: TAB/fixactfun.tex
\begin{table}[t]
	\begin{center}
		\begin{tabular}{|c|c|c|}
			\hline
			\textbf{Name} &\textbf{Expression} & \textbf{Range}\\
			\hline
			Identity & $\text{id}(a)=a$ & $(-\infty,+\infty)$\\
			Step (Heavyside) & 	$Th_{\geq 0}(a)=\begin{cases}
			0 & \text{if } a<0\\
			1&\text{otherwise}
			\end{cases}$ & $\{0,1\}$\\
			Bipolar & $B(a)=\begin{cases}
			-1 & \text{if } a<0\\
			+1&\text{otherwise}
			\end{cases}$ & $\{-1,1\}$\\
			Sigmoid & $\sigma(a)=\frac{1}{1+e^{-a}}$ & $(0,1)$\\	
			Bipolar sigmoid & $\sigma_B(a)=\frac{1-e^{-a}}{1+e^{-a}}$ & $(-1,1)$\\
			Hyperbolic tangent & 		$\tanh(a)$ & $(-1,1)$\\
			Hard hyperbolic tangent & $\tanh_H(a)=\max\big(-1,\min (1,a)\big)$ & $[-1,1]$ \\
			Absolute value & $\text{abs}(a)=|a|$& $[0,+\infty)$ \\
			Cosine & $\cos(a)$ & $[-1,1]$\\
			\hline
		\end{tabular}
	\end{center}
	\caption{Some of the \textcolor{black}{most used fixed activation functions.}}
	\label{tab:act_fun}
\end{table}

%% file: learn_shape.tex
\section{Trainable Activation functions}
\label{sec:trsh}
The idea of using trainable activation functions is not new in the neural networks research field. 
Many studies were published on this subject as early as the 1990s (see, for example, \citep{piazza1992,piazza1993,guarnieri1995,chen1996}). 
In more recent years, the renewed interest in neural networks has led the research to consider again the hypothesis that trainable activation functions could improve the performance of neural networks. 

In this section we describe and analyze the main methods presented in the literature related to the activation functions that can be learned by data. Relying on their main characteristics, we can isolate two distinct families:

\begin{itemize}
    \item Parameterized standard activation functions.
    \item Activation functions based on ensemble methods.
\end{itemize}

With \textit{parameterized standard activation} functions we refer to all the functions with a shape very similar to a given fixed-shape function, but tuned by a set of trainable parameters; with \textit{ensemble methods} we refer to any technique merging different functions. 

In the following of this section we discuss these two families of functions.


\subsection{Parameterized standard activation functions}
\begin{figure}
    \centering
    \includegraphics[width=\textwidth]{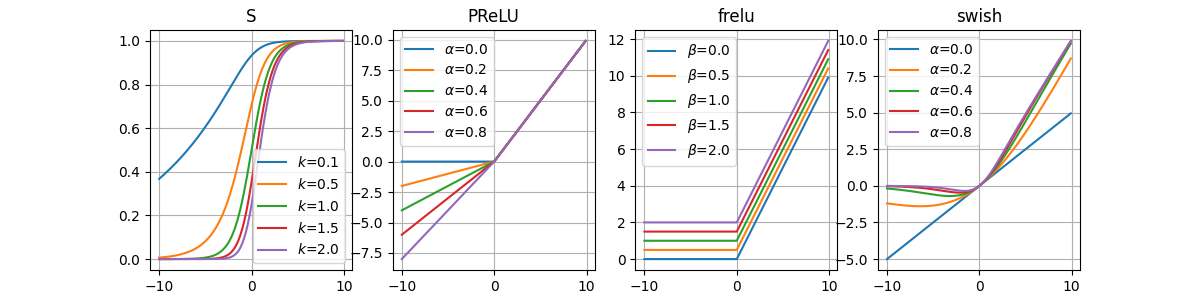}
     \includegraphics[width=\textwidth]{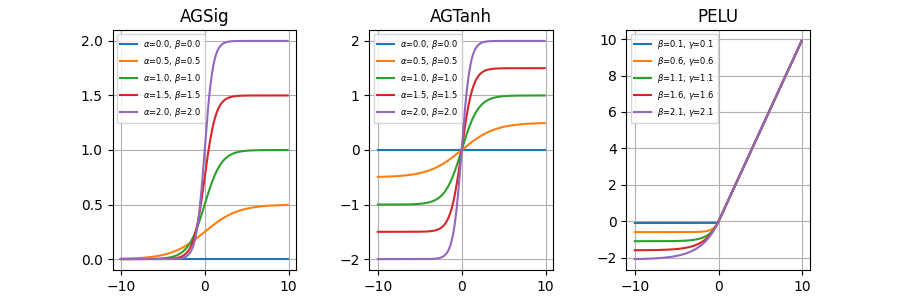}
    \caption{\textcolor{black}{Examples of parameterized standard activaton functions}}
    \label{fig:parameterized_af}
\end{figure}
As mentioned above, with the expression ``parameterized standard activation'' functions we refer to all the functions with a shape very similar to a given fixed-shape function, but having a set of trainable parameters that let this shape to be tuned. 
The addition of these parameters, therefore, requires changes, even minimal ones, in the learning process; for example, when using gradient-based methods, the partial derivatives of these new parameters are needed.

In the remainder of this subsection, we focus on the main functions which fall into this family, \textcolor{black}{some of which are represented in Figure \ref{fig:parameterized_af}}.

\paragraph{\textbf{Adjustable Generalized Sigmoid}}
A first attempt to have a trainable activation function was given in \citep{hu1992}. The proposed activation function was a generalization of the classic sigmoid  $\text{sig}(a)=\frac{1}{1+\exp({-a})}$ with the addition of two trainable parameters $\alpha,\beta$ to adjust the function shape, i.e.:
\begin{equation*}
    \text{AGSig}(a)=\frac{\alpha}{1+\exp(-\beta a)}
\end{equation*}
Both parameters are learned together using a gradient descent approach based on the backpropagation algorithm to compute the derivatives of the error function with respect to the network parameters.

\paragraph{\textbf{Sigmoidal selector}}
In \citep{singh2003} the following class of sigmoidal functions is proposed as:
\begin{equation*}
    \text{S}_k(a)=\Big(\frac{1}{1+\exp(-a)}\Big)^k
\end{equation*}
These functions are parameterized by a value $k \in (0,+\infty)$. 
In \citep{chandra2004} the  parameter $k$ is learned (and so the effective function is selected) together with the other network parameters by the gradient descent and backpropagation algorithms. 

\paragraph{\textbf{Adjustable Generalized Hyperbolic Tangent}}
Proposed in \citep{chen1996}, this activation function generalizes the classic hyperbolic tangent function $\tanh(a)=\frac{1 - e^{-2a}} {1 + e^{-2a}}$ introducing two trainable parameters $\alpha, \beta$:
\begin{equation*}
    \text{AGTanh}(a)=\frac{\alpha \big(1-\exp(-\beta a)\big)}{1+\exp(-\beta a)}
\end{equation*}
In this function, $\alpha$ adjusts the saturation level, while $\beta$ controls the slope. 
These two parameters are learned together with the network weights using the classic gradient descent algorithm combined with back-propagation, initializing all the values randomly. 
In \citep{yamada1992,yamada1992_2} a similar activation function was proposed, with the main difference that it included only one trainable parameter \textcolor{black}{controlling} the slope.

\paragraph{\textbf{Parametric ReLU}}
\citep{he2015} introduced another ReLU-like function which partially learns its shape from the training set, in fact it can modify the negative part using a parameter $\alpha$; the function Parametric ReLU, can be defined as:
\begin{equation*}
    \text{PReLU}(a) = 
    \begin{cases}
    a & \text{ if }a > 0 \\
    \alpha\cdot a & \text{otherwise.}
    \end{cases}
\end{equation*}
The additional parameter $\alpha$ is learned jointly with the whole model using classical gradient-based methods with backpropagation without weight decay to avoid pushing $\alpha$ to zero during the training.
From a computational point of view, the additional parameter appears negligible if compared to the total number of network parameters when an $\alpha$-sharing policy is used. 
Empirical experiments show that the magnitude of $\alpha$ rarely is larger than $1$, although no constraints on its range are applied.
However, the resulting function remains basically a modified version of the ReLU function that can change its shape just in the negative part.

\paragraph{\textbf{Parametric ELU}}
\citep{trottier2017} tries to eliminate the need to manually set the $\alpha$ parameter of ELU unit by proposing an alternative version based on two trainable parameters, i.e.
\begin{equation*}
    \text{PELU}(a) = 
    \begin{cases}
    \frac{\beta}{\gamma} a & \text{ if }a \geq 0 \\
    \beta\cdot (\exp (\frac{a}{\gamma})-1 ) & \text{otherwise.}
    \end{cases}
\end{equation*}
where $\beta,\gamma >0$ control the function shape and are learned together with the other network parameters using some optimization gradient-based method. 

\paragraph{\textbf{Flexible ReLU}}
Authors in \citep{qiu2018} propose the following function:
\begin{equation*}
    \text{frelu}(a)= ReLU(a + \alpha) + \beta 
\end{equation*}
with $\alpha,\beta$ learned by data. This is done to capture the negative information which is lost with the classic ReLU function and to provide the zero-like property \citep{clevert2015}. Considering that the value of $a$ is a weighted sum of inputs and bias, the $\alpha$ parameter can be viewed as part of the function input, so the authors reduced the function to
\begin{equation*}
    \text{frelu}(a)= ReLU(a) + \beta.
\end{equation*}

\paragraph{\textbf{Swish}}
\label{sec:swish}

In \citep{ramachandran2017} the authors propose a search technique for activation functions. In a nutshell, a  set of candidate activation functions is built by combining functions belonging to a predefined set of basis activation functions. For each candidate activation function, a network which uses generated function is trained on some task to evaluate the performance. Between all the tested functions, the best one results to be:
\begin{equation*}
    \text{Swish}(a)= a  \cdot \text{sig}(\alpha \cdot a) 
\end{equation*}
where $\alpha$ is a trainable parameter. it is worth pointing out that for $\alpha \to +\infty$, Swish behaves as ReLU, while, if $\alpha = 1$, it becomes equal to SiLU \citep{elfwing2018}. 

\subsubsection*{\textbf{Discussion}}
The activation functions proposed in this section share the peculiarity to be based on standard fixed activation functions whose shape is tuned using one or more trainable parameters.
However, the trained activation function shape turns out to be very similar to its corresponding non-trainable version, with the result of a poor increase of its expressiveness.
For example, see Figure \ref{fig:parameterized_af}, AGSig (AGTanh) function results to be the sigmoid (Tanh) function with smoothness and amplitude tuned by the $\alpha$ and $\beta$ parameters. Similarly, Swish can be viewed as a parameterized SiLU/ReLU variation whose final shape is learned to have a good trade-off between these two functions. In the end, the general function shape remains substantially bounded to assume the basic function(s) shape on which it has been built. 


However, being the most significant part of these functions just a weighted output of the respective weighted input of a fixed activation function, one can notice that it is possible to model each of them by a simple shallow neural subnetwork composed of few neurons. 
As an illustrative example, consider a neuron $n$ with the Swish activation function and weighted sum of inputs $a$, $n$'s behaviour is equivalent to a feed forward subnetwork composed of three neurons $n_1,n_2,n_3$ with respective weighted sums of inputs $a_1,a_2,a_3$ (see Figure \ref{fig:swishnn}): $n_1$ corresponds to the neuron $n$ equipped with the identity function instead of Swish,  $n_2$ is a neuron equipped with fixed activation function $f_2(a_2) = a_2\sigma(a_2)$ without bias value, and the last neuron $n_3$ has, again, the identity as activation function and no bias associated. 
\textcolor{black}{With} the connection going from $n_1$ to $n_2$ a weight with value $\alpha$ is associated, learned by data together with the other network parameters. The weight ${\alpha}$ assumes the role of the parameter $\alpha$ of the Swish function, while the connection between $n_2$ and $n_3$ has to be constrained to assume the value $\frac{1}{{\alpha}}$. 
In fact, the weighted sum fo the inputs of $n_2$ is $a_2={\alpha} a_1$ and the $n_3$ final output results to be $\frac{1}{{\alpha}} a_2 \cdot \sigma(a_2)=\frac{1}{{\alpha}}{\alpha} \cdot a_1 \cdot \sigma({\alpha} \cdot a_1)=a_1 \cdot \sigma({\alpha} a_1)$. Since $a_1=a$, $n_3$ output will be equal to $Swish(a)$. 

Similar consideration can be made for other trainable activation functions, some of which we report in \textcolor{black}{Figures~\ref{fig:srelunn} and \ref{fig:nnform}.}
\input{FIGURE/SWISH.tex}

Another aspect that is worth to stress is that for all the above mentioned trainable activation functions a gradient descent method based on backpropagation was used to train all the parameters of the networks; however, to train the activation function parameters the standard backpropagation formulas have to be suitable and specifically adapted to each trainable activation function.

\subsection{Functions based on ensemble methods} 
With the expression ``ensemble methods'' we refer to any technique merging together different functions. Basically, each of these techniques uses:
\begin{itemize}
    \item a \textit{set of basis functions}, which can contain fixed-shape functions or trainable functions or both;
    \item a \textit{combination model},  which defines how the basis functions are combined together.
\end{itemize}


As a significant example of this type of approach, in \citep{ramachandran2017} a method to investigate activation functions  built as compositions of several unary and binary functions is proposed, together with a search technique which works in these spaces. 
In this framework, a set of basis functions is provided (e.g., $\{\sin(a), \max(a,0),...\}$) together with the number of inputs (unary, binary,...) of each function. 
In this case, the combination model is the structure that the final activation function must follow in terms of function compositions (for example, the final output is given by the output of a binary function which has as input \textcolor{black}{two unary functions} and so on). 
The authors state that \textcolor{black}{several models} (that they called \textit{search spaces}) have been tried using different combinations of unary and binary functions. 
Among all the models investigated, we mention \textit{Swish} function, which has been discussed in \ref{sec:swish}. 
A similar methodology was made in \citep{basirat2019}, but using a genetic algorithm to learn the best activation function.

However, currently, a significant subset of works based on ensemble methods privileges the use of unary functions combined linearly.
In other terms, it is possible to identify a relevant sub-family of works sharing the common feature that the activation functions proposed can be expressed in terms of a linear combination of one-to-one functions, that is constraining the basis functions set to be composed of only functions of the type $g \in \mathbb{R}$, and the combination model to be a linear model.
In the remainder of this subsection we focus on this particular subclass of ensemble functions, highlighting how these can be easily expressed as sub-networks, and easily integrated into the model without constructing ad-hoc neurons, and without the need of modifying the learning algorithm, but only using classical feed-forward neural networks layers.

\subsubsection{Linear combination of one-to-one functions}
\label{sec:LinearCombinationOneToOneFunctions}
In this section we will group all the works which present trainable activation functions which can be expressed as a combination of several one-to-one functions.  
In a nutshell, the resulting trainable activation function can be reduced to $f(a)=\sum\limits_{i=1}^h \alpha_i\cdot g_i(a)$ where $g_1,g_2,\dots,g_h$ are one-to-one mapping functions, that is $g_i: \mathbb{R} \to \mathbb{R}, \ \forall 1 \leq i \leq h$ and $\alpha_i$ are weights associated with the functions, usually learned by data. 
Thus, in this type of approach the combination model is the linear combination and the basis functions are one-to-one mapping functions.

All the works reviewed in this section can be reduced to this general form. 
However, we distinguish at least two different ways in which they are presented originally: the former is based on the fact that the linear combination of functions can be expressed as a neural network; in the latter, the trainable activation \textcolor{black}{functions are expressed} using an analytical form. 
However, we will see that also for several of these functions a simple equivalent formulation in terms of subnetworks exists. 
In the next part of this Section, we will review the main works, dividing them according to the way they are originally presented.

\paragraph{\textbf{Adaptive Activation functions}}
In \citep{qian2018} the authors present, in a probabilistic and hierarchical context, different mixtures of the ELU and ReLU functions able to obtain a final activation function learned from data. 
The authors propose the following activation functions:
\begin{itemize}
    \item Mixed activation: $\Phi_{M}(a)=p\cdot \text{LReLU}(a)+(1-p)\cdot \text{ELU}(a)$ with $p\in [0,1]$ learned from data;
    \item Gated activation: $\Phi_{G}(a)=\text{sig}(\beta a)  \cdot \text{LReLU}(a)+(1-\text{sig}(\beta a))\cdot \text{ELU}(a)$ with $\text{sig}(\cdot)$ the sigmoid function and $\beta$ learned from data;
    \item Hierarchical activation: this function is composed of a three-level subnetwork, where each input unit $u$ is connected to $n$ units, and every pair of nodes are combined similarly as in gated activation (with the substitution of the ELU functions with PELU and ReLU with PReLU). At the same time, the last layer takes the maximum of the middle-level unit. So, the hierarchical organization can be formalized as follows:
    \begin{equation*}
     \begin{cases}
        \phi_{l,i}^{(1)}(a)=\text{PReLU}(a) \text{ and } \phi_{r,i}^{(1)}(a)=\text{PELU}(a) \\
        \phi_i^{(2)}(a)=\text{sig}(\beta_i a)  \cdot \phi_{l,i}^{(1)}(a)+(1-\text{sig}(\beta_i a))\cdot \phi_{r,i}^{(1)}(a) \\
        \phi^{(3)}(a)=\max\limits_{i} \phi_i^{(2)}(a) 
    \end{cases}
     \end{equation*}
    where $\phi_{l,i}^{(1)}(a)=$ is for the units of the first level, with $i=1,2,\cdot, m$, $\phi_i^{(2)}(a)$ is used for the $i$-th unit of the second level, and $\phi^{(3)}(a)$ is for the therd level; the final activation function is $\Phi_H(a)=\phi_l^{(3)}(a)$.
\end{itemize}
    

\paragraph{\textbf{Variable Activation Function}}
In \citep{apicella2019} trainable activation functions are expressed in terms of sub-networks with only one hidden layer, relying on the consideration that a one-hidden layer neural network can approximate arbitrarily well any continuous functional mapping from one finite-dimensional space to another, enabling the resulting function to assume ``any'' shape. 
In a nutshell, the proposed activation is modelled as a non-linear activation function $f$ with a neuron with an Identity activation function which sends its output to a one-hidden-layer sub-network with just one output neuron having, in turn, an Identity as an output function. 

The proposed model can be formalized as 
\begin{equation}\label{eq:vaf}
VAF(a)= \sum\limits_{j=1}^k \beta_j g(\alpha_j a + \alpha_{0j})+\beta_0
\end{equation}
where $g$ is a fixed-shape activation function, $k$ a hyper-parameter that determines the number of hidden nodes of the subnetwork and $\alpha_j$, $\alpha_{0j}$, $\beta_j$,  $\beta_0$ are parameters learned from the data during the training process.

\paragraph{\textbf{Kernel-based Activation Function}}
\label{sec:kaf}
The activation function proposed in \citep{scardapane2018} is modelled in terms of a kernel expansion:
\begin{equation}
    KAF(a) = \sum\limits_{i=1}^D \alpha_ik(a,d_i)
\end{equation}
where ${\alpha_1,\alpha_2,\dots,\alpha_D}$ are the trainable parameters,$k$ a kernel function $k: \mathbb{R} \times \mathbb{R} \to \mathbb{R}$ and ${d_1,d_2,\dots,d_D}$ are the dictionary elements, sampled from the real line for simpleness. The choice of the kernel function is widely discussed in \citep{scardapane2018}. 
\textcolor{black}{Furthermore, in \citep{scardapane2018complex}, the authors show two different methods to make KAF adapt to be used also with Complex-valued Neural Networks \citep{nitta2013local}.}


\paragraph{\textbf{Adaptive Blending Unit}}
The work proposed in \citep{sutfeld2018} combines together a set of different functions in a linear way, that is:
\begin{equation*}
    \text{ABU}(a)= \sum\limits_{i=1}^k{\alpha_i \cdot {f_i(a)} }
\end{equation*}
with $( \alpha_0, \alpha_1, \alpha_2,\dots,  \alpha_k)$ parameters to learn and $(f_1(\cdot), f_2(\cdot),\dots,f_k(\cdot))$ a set of activation functions that is, in the original study,  composed of tanh, ELU, ReLU, id, and Swish. The $\alpha$ parameters are all initialized with $\frac{1}{k}$ and are constrained using four different normalization schemes.

\paragraph{\textbf{Adaptive Piecewise Linear Units}}
In \citep{agostinelli2014} the activation functions are modeled as a sum of hinge-shaped functions that results in a different piece-wise linear activation function for every neuron:
\begin{equation*}
    \text{APL}(a) = \max (0,a) + \sum\limits_{i =1}^k w_{k} \max (0,-a + b_{k})
\end{equation*}
where \textcolor{black}{$k$ is a} hyper-parameter and  $w_{k},b_{k}$ are parameters learned during the network training. \textcolor{black}{The authors show that APL can learn convex and non-convex functions. However,} the total overhead in terms of number of parameters to learn compared with a classic NN with $n$ units is $2\cdot k \cdot n$, so the number of parameters increases with the number of hidden units and, for a large input, the learned function tends to behave as a ReLU function, reducing the expressiveness of the learned activation functions.
In the experiments reported in \citep{agostinelli2014} the value of $k$  was determined using a validation process,
while the $w$ and $b$ parameters were regularized with an $L_2$ penalty, so that the optimizer can avoid numerical instability leaving out too large values for the parameters.

\paragraph{\textbf{Harmon \& Klabjan Activation Ensembles}}
Some studies try to define activation functions using  different available activation functions rather than creating an entirely new function. 
For example, in \citep{harmon2017} the authors allow the network to choose the best activation function from a predefined set $F=\{f^{(1)},f^{(2)},\dots,f^{(k)}\}$, or some combination of those. 
Differently from Maxout (see Section \ref{sec:maxout}), the activation functions are combined together instead of simply taking the function with the maximum value. 

The activation function proposed by \citep{harmon2017} works on single mini-batch, i.e. its input is tuple $\vec{a}^{(u)}=(a^{(u)}_1,a^{(u)}_2,\dots, a^{(u)}_B)$ where every $a^{(u)}_i,\ 1\leq i \leq B$  refers to the unit $u$ on the $i$-th element of the mini-batch. 
The proposed activation function is based on a sum of normalized functions weighted by a set of learned weights; the resulting activation function $\Phi(a)$ of $\vec{a}_b^{(u)}$ has the form:
\begin{equation*}
\Phi^{(u)}(a_b^{(u)})= \sum\limits_{j=1}^{k} \alpha_ u^{j}  \big( \eta^{(j)} g^{j}(a_b^{(u)}+\delta^{(j)}) \big)
\end{equation*}
where $\alpha_u^j$ is a weight value for the $u$-th unit and the $j$-th function, $\eta^{(j)}$ and $\delta^{(j)}$ are inserted to allow the network choosing to leave the activation in its original state if the performance is particularly good and 
\begin{equation*}
    g^{j}(a_b^{(u)}) = \frac{f^{j}(a^{(u)}_b)-\min\limits_i f^{j}( a^{(u)}_i )}{\max\limits_i f^{j}(a^{(u)}_i )- \min\limits_i f^{j}( a^{(u)}_i )+\epsilon}
\end{equation*}
where $\epsilon$ is a small number. 
The authors emphasize that, during the experiments, many neurons favored the ReLU function since the respective $a^{(u)}_i$ had greater magnitude compared with the others.
The learning of the $\alpha$ values was done in terms of an optimization problem with the additional constraint that $\alpha$ values must be non-zero and sum to one to limit the magnitude. So, the approach proposed by \citep{harmon2017} seems to require additional computational costs due to the resolution of a new optimization problem together with the standard network learning procedure.

\paragraph{\textbf{S-Shaped ReLU}}
Taking inspiration from the Webner-Fechner law \citep{fechner1966} and Stevens law \citep{stevens1957}, the authors of \citep{jin2016} designed an activation $S$-shape function determined by three linear functions: 
\begin{equation*}
    \text{SReLU}(a)=\begin{cases}
    \beta_1 + \alpha_1(a-\beta_1) & \text{ if } a \leq \beta_1\\
    a  & \text{ if }  \beta_1 < a < \beta_2\\
    \beta_2 + \alpha_2(a-\beta_2) & \text{ if } a \geq \beta_2\\
    \end{cases}
\end{equation*}
where $b_1,w_1,b_2,w_2$ are parameters that can be learned together with the other network parameters. 
Also, in this case, the weight decay cannot be used during the learning because it tends to pull the parameters to zero.
SReLU can learn both convex and non-convex functions, differently from other trainable approaches like Maxout unit \citep{goodfellow2013} that can learn just convex function. 
Furthermore, this function can approximate also ReLU when $b_2 \geq 0, w_2=1, b_1=0, w_1=0$ or LReLU/PReLU when $b_2 \geq 0, w_2=1, b_1=0, w_1>0$.

\subsubsection*{\textbf{Discussion}}
As stated above, a linear combination of one-to-one functions is always expressible as a linear combination of one-to-one mappings. 
Due to this common property, several of them can be expressed in terms of feed-forward neural networks. 
Some of these functions, as VAF and Adaptive Activation Functions, are already modelled as sub-networks that can be integrated into the main architecture without changing the learning algorithm in the respective presentation works. 
The sub-network structure allows the function to be trained in a ``transparent'' way for the rest of the network and the chosen training algorithm. 
On the other hand, several of the remaining functions are analytically presented by the authors instead of using the neural network paradigm. However, for several of these functions we show that a natural equivalent formulation in terms of subnetworks exists, making these architectures not only easy to integrate into the main models, but also easier to study using the general rules of feed-forward neural networks. We report some of the equivalent models \textcolor{black}{in Figures~\ref{fig:swishnn},  \ref{fig:srelunn} and \ref{fig:nnform}.}  
For instance, the SReLU can be expressed as 
\begin{multline*}
    SReLU(a) = 
    -\alpha_1ReLU(\beta_1-a) + \beta_1Th_{\geq0}(\beta_1-a)+\\
    + \beta_1Th_{>0}\big(\beta_2Th_{>0}(\beta_2-a)-ReLU(\beta_2-a)-\beta_1\big)+\\+ReLU\big(\beta_2Th_{>0}(\beta_2-a)-ReLU(\beta_2-a)-\beta_1\big)+\\
    + \alpha_2ReLU(\beta_2-a) + \beta_2Th_{\geq0}(\beta_2-a)\\
\end{multline*}
where $Th_{>0}(x)=1-Th_{\geq 0}(-x)$ and $\beta_1\geq\beta_2$. So the SReLU functions (and all the others 1-to-1 functions) can be expressed as FFNN with constraints on the parameters and the inner activation functions.
\input{FIGURE/SSHAPED.tex}

\subsection{Outliers}
\label{sec:outliers}
In \citep{piazza1992} an attempt to model activation functions using a polynomial activation function with adaptable coefficients was proposed. 
For a given degree $k \in \mathbb{N}$, the relative polynomial function is
\begin{equation*}
    \text{AP}(a) = \sum\limits_{i=0}^k  \alpha_i \cdot a^{i} 
\end{equation*}
with $( \alpha_0, \alpha_1, \alpha_2,\dots,  \alpha_k)$ parameters to learn. These function parameters can be learned together with the network parameters using gradient descent with back-propagation. 
However, it must be taken into account that networks with polynomial activation functions are not universal approximator, as shown in \citep{stinchcombe1990}.

\textcolor{black}{In \cite{trentin2001networks} the effects of learning the amplitude of the activation functions are investigated. The authors of \cite{castelli2014combination} associate to each learnable activation function (in the form of a MLP) a corresponding probabilistic measure which affects the activation function learning.} \citep{eisenach2016} tries to approximate an activation function using a Fourier expansion. This study uses a $2$-stage Stochastic Gradient Descent (SGD) algorithm to learn the parameters of the activation functions and of the network. \citep{urban2017} tries to learn the activation function using Gaussian processes while
\citep{goh2003} learns the amplitude of the activation functions in Recurrent Neural Networks \citep{cardot2011}.
\citep{ertuugrul2018} proposes two trainable activation functions using linear regression, but using network architectures different from the classic Feed-Forward Neural Networks (see for example \citep{chen2018,huang2015}).

\textcolor{black}{In the context of fuzzy activation functions \citep{jou1991fuzzy,karakose2004type,nguyen2017models}, in \citep{beke2019interval,beke2019learning} a neural unit based on Type-2 fuzzy logic is proposed, which is capable of representing simple or sophisticated activation functions through three hyperparameters defining the slopes in the negative and positive quadrants. These hyperparameters can be set as adjustable parameters which can be learned together with the other network ones.}

Other studies propose activation functions whose shape is computed using interpolation techniques. 
These techniques may need some additional input, depending on the technique used (for example a set of sampled points from a start function).

In \citep{guarnieri1995} the authors introduced the use of spline based activation functions whose shape can be learned by data using a set of $Q$ representative points. 
This method has been improved by \citep{vecci1998,scardapane2016}. 
More in detail, this technique tries to find a cubic spline to model the activation function sampling the control points from a sigmoid (as in \citep{guarnieri1995})  or from another function  (e.g. hyperbolic tangent, as in \citep{scardapane2016}) assuring universal approximation capability. 
The resulting function can be expressed as follows:
\begin{equation*}
    \text{SAF}(a)=\vec{u}^T B \vec{q}_{i:i+P}
\end{equation*}
where:
\begin{itemize}
    \item $i$ is the index of the closest knot; %
    \item $\vec{q}$ is the knots vector, with $\vec{q}_{i:i+P}:=(q_i, q_{i+1},\dots,q_{i+P})^T$, so the output is computed by spline interpolation over the closest knot and its $P$ right-most neighbors. supposing that the knots are uniformly spaced, i.e. $\vec{q}_{i+1} = \vec{q}_i + \Delta t$, for a fixed $\Delta t \in \mathbb{R}$, the normalized abscissa value can be computed as $u= \frac{a}{\Delta t} - \lfloor  \frac{a}{\Delta t} \rfloor$;
    \item $\vec{u}^T=(u^P,u^{P-1},\dots, u, 1 ) \in \mathbb{R}^{P+1}$ is the reference vector;
    \item $B \in \mathbb{R}^{(P+1)\times(P+1)}$ is the basis spline matrix. Different bases make different interpolation schemes; in \citep{vecci1998} the authors used the Catmull-Rom matrix \citep{smith1983}.
\end{itemize}
The $\vec{q}$ values are then adapted during the learning procedure, adding a regularization term on $\vec{q}$ to prevent the over-fitting. 
The regularization term results to be a very critical issue: while the authors of \citep{vecci1998} act on the $\Delta x$ value, in \citep{scardapane2016} the authors proposed to penalize changes in $\vec{q}$ compared with a ``good'' set of values, as for example the initial control points values, sampled from a standard NN activation function.

Based on a similar principle, in \citep{wang2018} the authors introduce Look-up Table Unit based on spline; in this work, the activation function is controlled by a look-up table containing a set of anchor-points that control the function shape. The look-up table idea is not new in trainable activation function field; a first example in this direction is found in \citep{piazza1993}, where a generic adaptive look-up table was addressed by the neuron linear combination and learned by data. The main difference with \citep{wang2018} is in the structure of the look-up table. It now returns the result of a spline interpolation instead of the raw number in the table.
More in detail, defining the set of anchor point as $A=\{(q_1,u_1),(q_2,u_2),\dots,(q_m,q_m)\}$ with $q_i=q_{1}+i \cdot \Delta t$ and $u_i$ the trainable parameters,  \citep{wang2018} proposes two different methods to generate the activation function; the first one by interpolation, which results in the function:
\begin{equation*}
\text{LuTU}(a) = \frac{1}{t} u_i (q_{i+1}-a) + u_{i+1}(a-q_{i}), \text{ if } q_{i} \leq a \leq q_{i+1}
\end{equation*}
and the second one using cosine smoothing:
\begin{equation*}
    \text{LuTU}(a) = \sum\limits_{i}^m u_i \cdot r(a-q_{i},\alpha t)
\end{equation*}
where $\alpha \in \mathbb{N}$ and
\begin{equation*}
    r(w,\tau) = \begin{cases}
    \frac{1}{2 \tau}(1+ \cos (\frac{\pi}{\tau}w) ) & \text{ if } -\tau \leq w \leq \tau\\
    0 & \text{otherwise.}
    \end{cases}
\end{equation*}
The method based on cosine smoothing was proposed to address the gradient instability suffered by the interpolation method.
This kind of approaches requires to set additional hyper-parameters like the function input domain, the number of anchor points and the space between them, the $\alpha$ value in \citep{wang2018} second approach or the spline type for \citep{vecci1998,scardapane2016} approaches. Beyond a robust mathematical formulation, these methods seem to require the tuning of several hyper-parameters.


\section{Trainable non-standard neuron definitions}
\label{sec:trainableTransferFunctions}
So far, we took care of activation functions in the classic meaning given by literature, i.e., a function $o(a)$ that builds the output of the neuron using as input the value returned by the internal transformation $\vec{w}\vec{x}+b$ made by the classic computational neuron model, as described in section \ref{sec:fixsh}. In the remainder of this section, we will review works which change the standard definition of neuron computation but are considered as neural network models with trainable activation functions in the literature.
In other terms, these functions can be considered as a different type of computational neuron unit compared with the original computational neuron model.

\paragraph{\textbf{Maxout unit}}
\label{sec:maxout}

\citep{goodfellow2013} was one of the first modern study that proposed a new kind of neural unit with a different output computation. 
The name \textit{Maxout} was given by the fact that the unit output is the max of a set of linear functions. 
Maxout units should not be considered simply activation function, since they use multiple weighted sums for every neuron instead of a single weighted sum $a=\vec{w}\vec{x}+b$ used with classical artificial neurons. 
More precisely, a Maxout unit makes a vector $\vec{a}=(a_1,a_2,\dots,a_k)$ 
with $\forall i \in \{1,k\}, \ a_i=\vec{w}^{(i)T} \vec{x}+b^{(i)}$ with $\vec{x} \in \mathbb{R}^{d}$ output given by the previous layer, $\{ \vec{w}^{(1)} \in \mathbb{R}^{d} ,\vec{w}^{(2)} \in \mathbb{R}^{d} ,\dots,\vec{w}^{(k)} \in \mathbb{R}^{ d} \}$, $\{b^{(1)} \in \mathbb{R}^m , b^{(2)} \in \mathbb{R}^m , \dots, b^{(k)} \in \mathbb{R}^m \}$.  ${w_i}$ and ${b_i}$ are the parameters to be learned.
In the end it returns
\begin{equation*}
    \text{Maxout}(\vec{a}) = \max\limits_{ 1 \leq j \leq k} \{ a_i  \}.
\end{equation*}
In other words, Maxout units take the maximum value over a subspace of $k$ trainable linear functions of the same input $\vec{x}$, obtaining a piece-wise linear approximator capable of approximating any convex function. 
The same model can be defined arranging all the $\vec{w}^{(i)}$ vectors as column of a single matrix $W \in \mathbb{R}^{d \times k}$ and all the $b^{(i)}$ scalars in a single vector $\vec{b} \in \mathbb{R}^{k}$, obtaining $\vec{a}=W^{T}\vec{x} + \vec{b} $.

To notice that, setting $k=2$ and 
$\vec{w}^{(1)}=0, b^{(1)}=0 $, Maxout becomes
\begin{equation*}
\begin{split}
\text{Maxout}(\vec{a})=& \max (\vec{w}^{(1)T}\vec{x}+b^{(1)},\vec{w}^{(2)T}\vec{x}+b^{(2)})=\\
&\max (0, \vec{w}^{(2)T}\vec{x}+b^{(2)})=\text{ReLU}(w^{(2)T}\vec{x}+b^{(2)}),
\end{split}
\end{equation*}
In a similar way, Maxout unit can be made equivalent to Leaky ReLU, so Maxout can be viewed as a generalization of classic rectifier-based units.
Being Maxout constituted by a set of feed-forward sub-networks, its parameters can be learned together with the whole network using classical gradient descent approaches.
By running a cross-validation experiment, in \citep{goodfellow2013} the authors  found that Maxout offers a clear improvement over ReLU units in terms of classification errors.
Despite the performance, this approach requires many new weights compared with a classic network based on ReLU and classical neural units, namely  $k$ times the number of parameters for every single neuron, significantly increasing the cost of the learning process.

\paragraph{\textbf{Multi-layer Maxout}}
The Maxout-based network is generalized in \citep{sun2018}. The authors adopt a function composition approach that they call Multi-layer Maxout Network (MMN) further increasing the number of parameters.
To limit the computational costs introduced, the authors proposed to replace just a portion of activation functions in traditional DNN with MMNs as a trade-off scheme between the accuracy and computing resources.

\paragraph{\textbf{Probabilistic Maxout}}
In \citep{springenberg2013} the authors describe Probout, a stochastic generalization of the Maxout unit trying to improve its invariance replacing the maximum operation in Maxout with a probabilistic sampling procedure, i.e.
\begin{equation*}
\text{Probout}(\vec{a})=a_i,  \text{ with } i \sim \text{Multinomial}(p_1,p_2,\dots,p_k)
\end{equation*}
where $p_i=\frac{\exp (\lambda a_i)}{\sum_{j=1}^k \exp (\lambda a_j)}$, and  $\lambda$ \textcolor{black}{is a hyperparameter.} 
The maximum operation substitution in Maxout arises from the observation that to use other operations could be useful to improve the performances. 
To notice that the Probout function reduces to Maxout for $\lambda \to +\infty$.

\paragraph{\textbf{NIN \& CIC}}
\label{sec:nin}
The authors of \citep{lin2013} proposed a trainable activation function designed for Convolutional Neural Networks. 
In this work, the activation functions of a convolution layer are replaced with a  full-connected multilayer perceptron. 
In the following of this paragraph, we indicate with:
\begin{itemize}
    \item $X \in \mathbb{R}^{h \times w \times c}$, the input of a CNN;
    \item $X_{ij}\in \mathbb{R}^{t \times t \times c}$ an input \textit{patch} of size $t$ centered in the position $ij$, that is the submatrix composed of all the elements of $X$ belonging to the rows $i-t/2, i-t/2+1, \dots, i+t/2$ and the columns $j-t/2, j-t/2+1, \dots, j+t/2$ and each channel $ 1,2,\dots, c$.
\end{itemize}

The Network in Network (NIN) architecture proposed by \citep{lin2013} is based on MLP sub-nets with $l$ layers nested into a CNN and applied to every patch $X_{ij}$. 
This operation results in a set of functions $f^{(i)}$ that can be expressed as:
\begin{align*}
  f^{(1)}_{i,j,c_1}&=\text{ReLU} (W^{(1)}_{c_1} X_{ij}+\vec{b}_{c_1}), \\
&\vdots\\
 f^{(l)}_{i,j,c_l}&=\text{ReLU} (W^{(l)}_{c_l} f^{(l-1)}_{i,j,*}+\vec{b}_{c_l}).
\end{align*}
where $l$ is the number of layers of the MLP and $c_i$ is the number of the input channels for $i=1$ or the number of the level filters for every $i>1$. We indicate with $f^{l-1}_{i,j,*}$ the vector composed of the functions output in the point $(x,j)$ for each channel, i.e. $f^{(l)}_{i,j,*}=(f^{(l)}_{i,j,1},f^{(l)}_{i,j,2}, \dots f^{(l)}_{i,j,c_l} )$.

So, this subnet can be viewed as an MLP with ReLU units and the output results to be a map constituted by the output of the last layer functions $f^{(l)}$.
Despite the good performances obtained, \textcolor{black}{these methods requires to learn a lot of extra parameters,} especially when $l$ is large.
NIN seems to move away from the concept of variable activation function of a single neuron because the MLP could have common connections with other nodes of the previous layers, in other words there are no constraints on the number of the output that the final layer of the MLP can have more than one output. 

Based on NIN, authors of \citep{pang2016} proposed Convolution in Convolution (CIC) which uses a sparse MLP instead of the classic full-connected MLP as activation function. 

\paragraph{\textbf{Batch-Normalized Maxout NIN}}
In \citep{chang2015} Maxout and NIN are combined togegher with Batch Normalization \citep{ioffe2015}. 
The proposed method replaces the ReLU functions present in NIN with Maxout to avoid the zero saturation and adds Batch Normalization to avoid the problems connected with changes in data distribution \citep{ioffe2015}. 
\textcolor{black}{
\paragraph{\textbf{LWTA}}
\label{sec:LWTA}
In \citep{srivastava2013compete} an interesting neural subnetwork based on the local competition between several neurons is presented. Relying on biological models of brain processes, a Local Winner-Take-All subnetwork (LWTA) consists in blocks of neurons whose outputs are fed in a \textit{competition/interaction} function that returns the final output of the block neurons.
Similarly to Maxout, given a set (block) of $k$ neurons, with activation given by $\ a_i=\vec{w}^{(i)T} \vec{x}, \forall i \in \{1,\dots, k\}$ with $\vec{x} \in \mathbb{R}^{d}$ output given by the previous layer and $\{ \vec{w}^{(1)} \in \mathbb{R}^{d} ,\vec{w}^{(2)} \in \mathbb{R}^{d} ,\dots,\vec{w}^{(k)} \in \mathbb{R}^{ d} \}$ parameters to be learned, the final output of a neuron $i$ is given by a function of all the neurons belonging to the same block of $i$, that is:
$$LWTA_i(\vec{a}) = \begin{cases}
a_i & \text{if } a_i=\max{(\vec{a})}\\
0   & \text{otherwise}
\end{cases}, \ \forall i \in \{1,\dots , k\}$$
that is, only the ``winner'' neuron will propagate its output to the next layer of the network, turning off the activations of the remaining neurons in the same block.
In this way, for a given input, only a subset of the network parameters is used. The authors  hypothesize that using different subsets of parameters for different inputs allows the architecture to learn more accurately.
}
\paragraph{\textbf{\textcolor{black}{Some considerations}}}
\textcolor{black}{We highlight that in some cases, the basic building blocks of a neural network are not the neurons, but groups of neurons, for example, the layers of a multi-layer feed-forward neural network (see,  \cite{eckle2019comparison}), and the activation function is expressed as a mapping $f: \mathbb{R}^k \longrightarrow \mathbb{R}^k$, where $k$ is the number neurons in the layer. However, this is a notational or computational simplification, and one can again consider the neuron as the primary building block and, consequently, the activation function as a function $f: \mathbb{R} \longrightarrow \mathbb{R}$.  Nevertheless, sometimes the input and activation functions of the basic building blocks differ substantially from the classical ones.  Maxout networks \citep{goodfellow2013} are a significant example. In the original paper, as above described, the authors propose a Maxout unit which, given an input $\vec{x} \in \mathbb{R}^d$, computes its output as $h(\vec{x}) = \max\limits_{i=1,\dots,k} \{a_i\}$, where $a_i= \vec{w}^{(i)T} \vec{x} +b^{(i)}$.  The input function of the maxout unit is a functional mapping $I: \mathbb{R}^d \to  \mathbb{R}^k$, and the activation function corresponds to the maximum operator. The authors themselves state that the maxout unit is equivalent to a standard one-hidden-layer feed-forward network (shallow network) with $k$ hidden neurons and one output neuron. The $k$ hidden neurons have input and activation functions corresponding to the standard weighted linear sum of the incoming input values and the identity function, respectively. The output neuron has the maximum as input function, and the identity as activation function. In the literature a Maxout network is often considered a network with trainable activation functions (see, for example, \cite{harmon2017,sun2018} ), however its activation function definition does not fit with the one we adopted in this paper. For this reason, in our taxonomy, we excluded the Maxout networks from the class of ``trainable activation function'', and we include this model into a specific ``trainable'' subclass of the ``Non-standard  neuron defitions'' class. 
\\Another significant example is Network in Network (NiN) \citep{lin2013}. This model is introduced in the context of Convolutional Neural Networks (CNN). The convolution filter in a CNN is a Generalized Linear Model (GLM), while in NIN the GLM is replaced by a more potent nonlinear function approximator \citep{lin2013}. In particular, the authors use as general nonlinear function a multilayer perceptron called ``mlpconv''. In the literature, this nonlinear filter is often interpreted as a single unit with trainable activation function (e.g., \cite{eisenach2016,harmon2017}). Also, in this case, this interpretation does not fit with the standard activation function definition which we adopted in this survey paper, thus in our taxonomy, we included this model in the same subclass of Maxout model.}

%% file: FIGURE/SWISH.tex
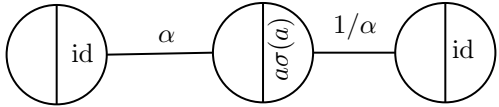
\begin{figure}
\tikzset{every picture/.style={line width=0.75pt}} 
\begin{center}
\begin{tikzpicture}[x=0.75pt,y=0.75pt,yscale=-1,xscale=1]

\draw   (134,126) .. controls (134,112.19) and (145.19,101) .. (159,101) .. controls (172.81,101) and (184,112.19) .. (184,126) .. controls (184,139.81) and (172.81,151) .. (159,151) .. controls (145.19,151) and (134,139.81) .. (134,126) -- cycle ;
\draw    (159,101) -- (159,151) ;

\draw   (237,125) .. controls (237,111.19) and (248.19,100) .. (262,100) .. controls (275.81,100) and (287,111.19) .. (287,125) .. controls (287,138.81) and (275.81,150) .. (262,150) .. controls (248.19,150) and (237,138.81) .. (237,125) -- cycle ;
\draw    (262,100) -- (262,150) ;

\draw    (184,126) -- (237,125) ;

\draw   (328,125) .. controls (328,111.19) and (339.19,100) .. (353,100) .. controls (366.81,100) and (378,111.19) .. (378,125) .. controls (378,138.81) and (366.81,150) .. (353,150) .. controls (339.19,150) and (328,138.81) .. (328,125) -- cycle ;
\draw    (353,100) -- (353,150) ;

\draw    (287,125) -- (328,125) ;

\draw (172,125) node  [align=left] {id};
\draw (271,123) node [rotate=-270] [align=left] {$\displaystyle a\sigma(a) $};
\draw (362,123) node  [align=left] {id};
\draw (214,117) node  [align=left] {$\displaystyle \alpha $};
\draw (308,112) node  [align=left] {$\displaystyle 1/\alpha $};

\end{tikzpicture}
\end{center}
\caption{A possible representation of the Swish activation function using just feed forward neural network layers. Connection with same labels are intended to be shared, while connection with numeric labels are intended to be fixed during the training. See text for further details. }
\label{fig:swishnn}
\end{figure}

%% file: FIGURE/SSHAPED.tex
\begin{center}
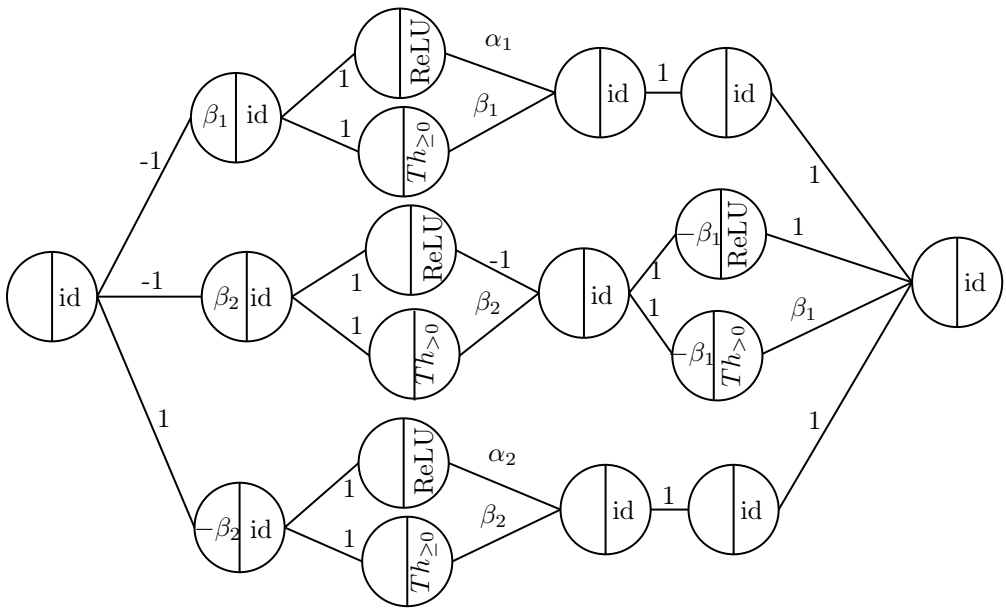
\begin{figure}[ht]
\tikzset{every picture/.style={line width=0.75pt}} 

\begin{tikzpicture}[x=0.75pt,y=0.75pt,yscale=-.9,xscale=.9]

\draw   (9,162) .. controls (9,148.19) and (20.19,137) .. (34,137) .. controls (47.81,137) and (59,148.19) .. (59,162) .. controls (59,175.81) and (47.81,187) .. (34,187) .. controls (20.19,187) and (9,175.81) .. (9,162) -- cycle ;
\draw    (34,137) -- (34,187) ;

\draw   (111,62) .. controls (111,48.19) and (122.19,37) .. (136,37) .. controls (149.81,37) and (161,48.19) .. (161,62) .. controls (161,75.81) and (149.81,87) .. (136,87) .. controls (122.19,87) and (111,75.81) .. (111,62) -- cycle ;
\draw    (136,37) -- (136,87) ;

\draw    (111,62) -- (59,162) ;

\draw    (252,26) -- (313,48) ;

\draw   (202,26) .. controls (202,12.19) and (213.19,1) .. (227,1) .. controls (240.81,1) and (252,12.19) .. (252,26) .. controls (252,39.81) and (240.81,51) .. (227,51) .. controls (213.19,51) and (202,39.81) .. (202,26) -- cycle ;
\draw    (227,1) -- (227,51) ;

\draw    (161,62) -- (202,26) ;

\draw   (204,81) .. controls (204,67.19) and (215.19,56) .. (229,56) .. controls (242.81,56) and (254,67.19) .. (254,81) .. controls (254,94.81) and (242.81,106) .. (229,106) .. controls (215.19,106) and (204,94.81) .. (204,81) -- cycle ;
\draw    (229,56) -- (229,106) ;

\draw    (161,62) -- (204,81) ;

\draw    (254,81) -- (313,48) ;

\draw   (304,160) .. controls (304,146.19) and (315.19,135) .. (329,135) .. controls (342.81,135) and (354,146.19) .. (354,160) .. controls (354,173.81) and (342.81,185) .. (329,185) .. controls (315.19,185) and (304,173.81) .. (304,160) -- cycle ;
\draw    (329,135) -- (329,185) ;

\draw   (117,162) .. controls (117,148.19) and (128.19,137) .. (142,137) .. controls (155.81,137) and (167,148.19) .. (167,162) .. controls (167,175.81) and (155.81,187) .. (142,187) .. controls (128.19,187) and (117,175.81) .. (117,162) -- cycle ;
\draw    (142,137) -- (142,187) ;

\draw    (117,162) -- (59,162) ;

\draw    (258,136) -- (304,160) ;

\draw   (208,136) .. controls (208,122.19) and (219.19,111) .. (233,111) .. controls (246.81,111) and (258,122.19) .. (258,136) .. controls (258,149.81) and (246.81,161) .. (233,161) .. controls (219.19,161) and (208,149.81) .. (208,136) -- cycle ;
\draw    (233,111) -- (233,161) ;

\draw    (167,162) -- (208,136) ;

\draw   (210,194) .. controls (210,180.19) and (221.19,169) .. (235,169) .. controls (248.81,169) and (260,180.19) .. (260,194) .. controls (260,207.81) and (248.81,219) .. (235,219) .. controls (221.19,219) and (210,207.81) .. (210,194) -- cycle ;

\draw    (235,170) -- (235,220) ;

\draw    (167,162) -- (210,195) ;

\draw    (260,195) -- (304,160) ;

\draw   (378,195) .. controls (378,181.19) and (389.19,170) .. (403,170) .. controls (416.81,170) and (428,181.19) .. (428,195) .. controls (428,208.81) and (416.81,220) .. (403,220) .. controls (389.19,220) and (378,208.81) .. (378,195) -- cycle ;
\draw   (380,127) .. controls (380,113.19) and (391.19,102) .. (405,102) .. controls (418.81,102) and (430,113.19) .. (430,127) .. controls (430,140.81) and (418.81,152) .. (405,152) .. controls (391.19,152) and (380,140.81) .. (380,127) -- cycle ;
\draw    (354,160) -- (378,194) ;

\draw    (354,160) -- (380,127) ;

\draw    (405,102) -- (405,152) ;

\draw    (401,170) -- (401,220) ;

\draw    (428,194) -- (511,154) ;

\draw    (430,127) -- (511,154) ;

\draw   (113,291) .. controls (113,277.19) and (124.19,266) .. (138,266) .. controls (151.81,266) and (163,277.19) .. (163,291) .. controls (163,304.81) and (151.81,316) .. (138,316) .. controls (124.19,316) and (113,304.81) .. (113,291) -- cycle ;
\draw    (138,266) -- (138,316) ;

\draw    (113,291) -- (59,162) ;

\draw    (254,255) -- (316,281) ;

\draw   (204,255) .. controls (204,241.19) and (215.19,230) .. (229,230) .. controls (242.81,230) and (254,241.19) .. (254,255) .. controls (254,268.81) and (242.81,280) .. (229,280) .. controls (215.19,280) and (204,268.81) .. (204,255) -- cycle ;
\draw    (229,230) -- (229,280) ;

\draw    (163,291) -- (204,255) ;

\draw   (206,310) .. controls (206,296.19) and (217.19,285) .. (231,285) .. controls (244.81,285) and (256,296.19) .. (256,310) .. controls (256,323.81) and (244.81,335) .. (231,335) .. controls (217.19,335) and (206,323.81) .. (206,310) -- cycle ;
\draw    (231,285) -- (231,335) ;

\draw    (163,291) -- (206,310) ;

\draw    (256,310) -- (316,281) ;

\draw   (313,48) .. controls (313,34.19) and (324.19,23) .. (338,23) .. controls (351.81,23) and (363,34.19) .. (363,48) .. controls (363,61.81) and (351.81,73) .. (338,73) .. controls (324.19,73) and (313,61.81) .. (313,48) -- cycle ;
\draw    (338,23) -- (338,73) ;

\draw   (316,281) .. controls (316,267.19) and (327.19,256) .. (341,256) .. controls (354.81,256) and (366,267.19) .. (366,281) .. controls (366,294.81) and (354.81,306) .. (341,306) .. controls (327.19,306) and (316,294.81) .. (316,281) -- cycle ;
\draw    (341,256) -- (341,306) ;

\draw   (383,48) .. controls (383,34.19) and (394.19,23) .. (408,23) .. controls (421.81,23) and (433,34.19) .. (433,48) .. controls (433,61.81) and (421.81,73) .. (408,73) .. controls (394.19,73) and (383,61.81) .. (383,48) -- cycle ;
\draw    (408,23) -- (408,73) ;

\draw   (387,281) .. controls (387,267.19) and (398.19,256) .. (412,256) .. controls (425.81,256) and (437,267.19) .. (437,281) .. controls (437,294.81) and (425.81,306) .. (412,306) .. controls (398.19,306) and (387,294.81) .. (387,281) -- cycle ;
\draw    (412,256) -- (412,306) ;

\draw   (511,154) .. controls (511,140.19) and (522.19,129) .. (536,129) .. controls (549.81,129) and (561,140.19) .. (561,154) .. controls (561,167.81) and (549.81,179) .. (536,179) .. controls (522.19,179) and (511,167.81) .. (511,154) -- cycle ;
\draw    (536,129) -- (536,179) ;

\draw    (363,48) -- (383,48) ;

\draw    (366,281) -- (387,281) ;

\draw    (433,48) -- (511,154) ;

\draw    (437,281) -- (511,154) ;

\draw (89,86) node  [align=left] {\mbox{-}1};
\draw (238,25) node [rotate=-270] [align=left] {ReLU};
\draw (125,61) node  [align=left] {$\displaystyle \beta_{1}$};
\draw (282,20) node  [align=left] {$\displaystyle \alpha_{1}$};
\draw (196,40) node  [align=left] {1};
\draw (197,68) node  [align=left] {1};
\draw (275,55) node  [align=left] {$\displaystyle \beta_{1}$};
\draw (238,81) node [rotate=-270] [align=left] {$\displaystyle Th_{\geq 0}$};
\draw (339,161) node  [align=left] {id};
\draw (131,163) node  [align=left] {$\displaystyle \beta_{2}$};
\draw (90,153) node  [align=left] {\mbox{-}1};
\draw (245,135) node [rotate=-270] [align=left] {ReLU};
\draw (283,141) node  [align=left] {\mbox{-}1};
\draw (203,154) node  [align=left] {1};
\draw (203,178) node  [align=left] {1};
\draw (276,166) node  [align=left] {$\displaystyle \beta _{2}$};
\draw (242,194) node [rotate=-270] [align=left] {$\displaystyle Th_{>0}$};
\draw (412,195) node [rotate=-270] [align=left] {$\displaystyle Th_{>0}$};
\draw (414,127) node [rotate=-270] [align=left] {ReLU};
\draw (451,169) node  [align=left] {$\displaystyle \beta_{1}$};
\draw (448,123) node  [align=left] {1};
\draw (393,127) node  [align=left] {$\displaystyle -\beta_{1}$};
\draw (389,194) node  [align=left] {$\displaystyle -\beta_{1}$};
\draw (240,254) node [rotate=-270] [align=left] {ReLU};
\draw (126,292) node  [align=left] {$\displaystyle -\beta_{2}$};
\draw (284,252) node  [align=left] {$\displaystyle \alpha_{2}$};
\draw (199,269) node  [align=left] {1};
\draw (199,297) node  [align=left] {1};
\draw (279,285) node  [align=left] {$\displaystyle \beta_{2}$};
\draw (240,310) node [rotate=-270] [align=left] {$\displaystyle Th_{\geq 0}$};
\draw (96,230) node  [align=left] {1};
\draw (369,147.5) node  [align=left] {1};
\draw (368,168) node  [align=left] {1};
\draw (148,60) node  [align=left] {id};
\draw (150,162) node  [align=left] {id};
\draw (150,292) node  [align=left] {id};
\draw (44,162) node  [align=left] {id};
\draw (348,48) node  [align=left] {id};
\draw (351,281) node  [align=left] {id};
\draw (418,48) node  [align=left] {id};
\draw (422,281) node  [align=left] {id};
\draw (546,154) node  [align=left] {id};
\draw (373,37) node  [align=left] {1};
\draw (376,273) node  [align=left] {1};
\draw (457,94) node  [align=left] {1};
\draw (457,231) node  [align=left] {1};

\end{tikzpicture}
\caption{A possible representation of the S-Shaped ReLU activation function using just feed forward neural network layers. Connection with same labels are intended to be shared, while connection with numeric labels are intended to be fixed during the training. See text for further details.}
\label{fig:srelunn}
\end{figure}
\end{center}

%% file: discussion.tex
\section{Performance and experimental architecture comparison}
\label{sec:disc}

All the reviewed studies highlight the improvements in terms of accuracy compared with using non-trainable activation functions in their model. 
However, making an exhaustive comparison between all the proposed approaches could be not very significant due to the differences in the experimental setup. 
Apart from works which make an explicit comparison with other trainable (and not) activation functions \citep{dasgupta1993,karlik2011,nwankpa2018}, \textcolor{black}{a complete comparison among all the existing architectures is not straightforward, even if the experiments are usually performed on standard and shared datasets,} as SVHN \citep{netzer2011}, MNIST \citep{lecun1998}, CIFAR10, CIFAR100 \citep{krizhevsky2009}, ImageNet \citep{ILSVRC2015}. 
This is due to different choices of neural networks architectures, learning algorithms and hyper-parameters in experimental setups. \textcolor{black}{With this in mind, in order to sketch an overview on this comparison,} in Table~\ref{tab:setup} we show the different architectures and datasets used in the main studies.
However, also for studies that use the same architectures, small changes to integrate the proposed model into the setup architectures may have been necessary.

Nevertheless, several works on trainable activation functions proposed in literature report improvements in comparison with equivalent architectures equipped with classical fixed-shape activation functions as ReLU or sigmoid or other trainable models.
Usually, this is shown through a comparison between identical setup architectures but different activation functions (see Table \ref{tab:accfun}). 
However, it is not clear if the performance improvements are due to the learning capability or are simply a consequence of the increased complexity of the final setup architecture. 

\input{FIGURE/nnform.tex}
\input{TAB/setup.tex}

%% file: FIGURE/nnform.tex
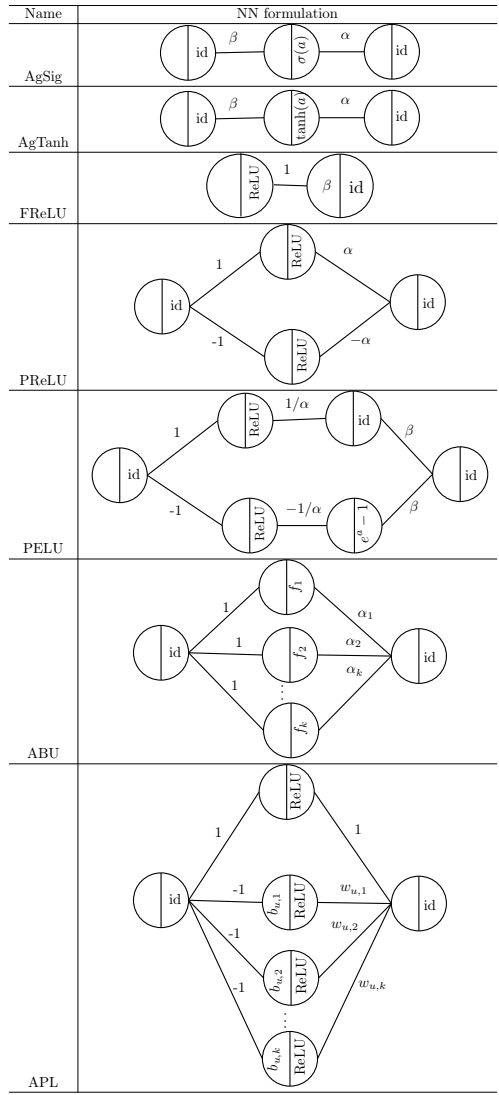
\begin{figure}[t]
\begin{center}
\scalebox{0.55}{
\begin{tabular}{c|c}
     \hline
     Name & NN formulation\\
     \hline
     AgSig &\input{FIGURE/AGSIG.tex}\\
     \hline
     AgTanh & \input{FIGURE/AGTANH.tex}\\
     \hline
     FReLU & \input{FIGURE/FRELU.tex}\\
     \hline
     PReLU & \input{FIGURE/PRELU.tex}\\
     \hline
     PELU & \input{FIGURE/PELU.tex}\\
     \hline
     ABU & \input{FIGURE/ABU.tex}\\
     \hline
     APL & \input{FIGURE/APL.tex}\\
     \hline
\end{tabular}
}
\end{center}
\caption{\textcolor{black}{Several trainable activation functions} represented as feed forward layers with constrained weights. The connections having the same labels have shared weights, that is the same value. The connection having numeric values as labels must have the same fixed value during the training.}
\label{fig:nnform}
\end{figure}
\FloatBarrier

%% file: FIGURE/AGSIG.tex
\tikzset{every picture/.style={line width=0.75pt}} 

\begin{tikzpicture}[x=0.75pt,y=0.75pt,yscale=-1,xscale=1]

\draw   (143,126) .. controls (143,112.19) and (154.19,101) .. (168,101) .. controls (181.81,101) and (193,112.19) .. (193,126) .. controls (193,139.81) and (181.81,151) .. (168,151) .. controls (154.19,151) and (143,139.81) .. (143,126) -- cycle ;
\draw    (168,101) -- (168,151) ;

\draw   (237,125) .. controls (237,111.19) and (248.19,100) .. (262,100) .. controls (275.81,100) and (287,111.19) .. (287,125) .. controls (287,138.81) and (275.81,150) .. (262,150) .. controls (248.19,150) and (237,138.81) .. (237,125) -- cycle ;
\draw    (262,100) -- (262,150) ;

\draw    (193,126) -- (237,125) ;

\draw   (328,125) .. controls (328,111.19) and (339.19,100) .. (353,100) .. controls (366.81,100) and (378,111.19) .. (378,125) .. controls (378,138.81) and (366.81,150) .. (353,150) .. controls (339.19,150) and (328,138.81) .. (328,125) -- cycle ;
\draw    (353,100) -- (353,150) ;

\draw    (287,125) -- (328,125) ;

\draw (181,125) node  [align=left] {id};
\draw (271,123) node [rotate=-270] [align=left] {$\displaystyle \sigma(a) $};
\draw (362,123) node  [align=left] {id};
\draw (311,110) node  [align=left] {$\displaystyle \alpha $};
\draw (209,113) node  [align=left] {$\displaystyle \beta $};

\end{tikzpicture}

%% file: FIGURE/AGTANH.tex
\tikzset{every picture/.style={line width=0.75pt}} 

\begin{tikzpicture}[x=0.75pt,y=0.75pt,yscale=-1,xscale=1]

\draw   (143,126) .. controls (143,112.19) and (154.19,101) .. (168,101) .. controls (181.81,101) and (193,112.19) .. (193,126) .. controls (193,139.81) and (181.81,151) .. (168,151) .. controls (154.19,151) and (143,139.81) .. (143,126) -- cycle ;
\draw    (168,101) -- (168,151) ;

\draw   (237,125) .. controls (237,111.19) and (248.19,100) .. (262,100) .. controls (275.81,100) and (287,111.19) .. (287,125) .. controls (287,138.81) and (275.81,150) .. (262,150) .. controls (248.19,150) and (237,138.81) .. (237,125) -- cycle ;
\draw    (262,100) -- (262,150) ;

\draw    (193,126) -- (237,125) ;

\draw   (328,125) .. controls (328,111.19) and (339.19,100) .. (353,100) .. controls (366.81,100) and (378,111.19) .. (378,125) .. controls (378,138.81) and (366.81,150) .. (353,150) .. controls (339.19,150) and (328,138.81) .. (328,125) -- cycle ;
\draw    (353,100) -- (353,150) ;

\draw    (287,125) -- (328,125) ;

\draw (181,125) node  [align=left] {id};
\draw (271,123) node [rotate=-270] [align=left] {$\displaystyle \tanh(a)$};
\draw (362,123) node  [align=left] {id};
\draw (311,110) node  [align=left] {$\displaystyle \alpha $};
\draw (209,113) node  [align=left] {$\displaystyle \beta $};

\end{tikzpicture}

%% file: FIGURE/FRELU.tex
\tikzset{every picture/.style={line width=0.75pt}} 

\begin{tikzpicture}[x=0.75pt,y=0.75pt,yscale=-1,xscale=1]

\draw   (70,65) .. controls (70,48.98) and (83.43,36) .. (100,36) .. controls (116.57,36) and (130,48.98) .. (130,65) .. controls (130,81.02) and (116.57,94) .. (100,94) .. controls (83.43,94) and (70,81.02) .. (70,65) -- cycle ;
\draw    (101,35) -- (101,93) ;

\draw   (161,65) .. controls (161,48.98) and (174.43,36) .. (191,36) .. controls (207.57,36) and (221,48.98) .. (221,65) .. controls (221,81.02) and (207.57,94) .. (191,94) .. controls (174.43,94) and (161,81.02) .. (161,65) -- cycle ;
\draw    (191,36) -- (191,94) ;

\draw    (131,64) -- (161,65) ;

\draw (113,62) node [rotate=-270] [align=left] {\small ReLU};
\draw (179,65) node  [align=left] {$\displaystyle \beta $};
\draw (143,50) node  [align=left] {1};
\draw (206,66) node [scale=1.2] [align=left] {id};

\end{tikzpicture}

%% file: FIGURE/PRELU.tex
\tikzset{every picture/.style={line width=0.75pt}} 

\begin{tikzpicture}[x=0.75pt,y=0.75pt,yscale=-1,xscale=1]

\draw   (248,66) .. controls (248,52.19) and (259.19,41) .. (273,41) .. controls (286.81,41) and (298,52.19) .. (298,66) .. controls (298,79.81) and (286.81,91) .. (273,91) .. controls (259.19,91) and (248,79.81) .. (248,66) -- cycle ;
\draw   (252,162) .. controls (252,148.19) and (263.19,137) .. (277,137) .. controls (290.81,137) and (302,148.19) .. (302,162) .. controls (302,175.81) and (290.81,187) .. (277,187) .. controls (263.19,187) and (252,175.81) .. (252,162) -- cycle ;
\draw    (277,137) -- (277,187) ;

\draw   (366,112) .. controls (366,98.19) and (377.19,87) .. (391,87) .. controls (404.81,87) and (416,98.19) .. (416,112) .. controls (416,125.81) and (404.81,137) .. (391,137) .. controls (377.19,137) and (366,125.81) .. (366,112) -- cycle ;
\draw   (134,116) .. controls (134,102.19) and (145.19,91) .. (159,91) .. controls (172.81,91) and (184,102.19) .. (184,116) .. controls (184,129.81) and (172.81,141) .. (159,141) .. controls (145.19,141) and (134,129.81) .. (134,116) -- cycle ;
\draw    (159,91) -- (159,141) ;

\draw    (273,41) -- (273,91) ;

\draw    (391,87) -- (391,137) ;

\draw    (184,116) -- (248,66) ;

\draw    (184,116) -- (252,162) ;

\draw    (298,66) -- (366,112) ;

\draw    (302,162) -- (366,112) ;

\draw (401,113) node  [align=left] {id};
\draw (172,115) node  [align=left] {id};
\draw (281,65) node [rotate=-270] [align=left] {{\small ReLU}};
\draw (286,160) node [rotate=-270] [align=left] {{\small ReLU}};
\draw (211,77) node  [align=left] {1};
\draw (210,148) node  [align=left] {\mbox{-}1};
\draw (328,65) node  [align=left] {$\displaystyle \alpha $};
\draw (339,147) node  [align=left] {$\displaystyle -\alpha $};

\end{tikzpicture}

%% file: FIGURE/PELU.tex
\tikzset{every picture/.style={line width=0.75pt}} 

\begin{tikzpicture}[x=0.75pt,y=0.75pt,yscale=-1,xscale=1]

\draw   (248,66) .. controls (248,52.19) and (259.19,41) .. (273,41) .. controls (286.81,41) and (298,52.19) .. (298,66) .. controls (298,79.81) and (286.81,91) .. (273,91) .. controls (259.19,91) and (248,79.81) .. (248,66) -- cycle ;
\draw   (252,162) .. controls (252,148.19) and (263.19,137) .. (277,137) .. controls (290.81,137) and (302,148.19) .. (302,162) .. controls (302,175.81) and (290.81,187) .. (277,187) .. controls (263.19,187) and (252,175.81) .. (252,162) -- cycle ;
\draw    (277,137) -- (277,187) ;

\draw   (443,115) .. controls (443,101.19) and (454.19,90) .. (468,90) .. controls (481.81,90) and (493,101.19) .. (493,115) .. controls (493,128.81) and (481.81,140) .. (468,140) .. controls (454.19,140) and (443,128.81) .. (443,115) -- cycle ;
\draw   (134,116) .. controls (134,102.19) and (145.19,91) .. (159,91) .. controls (172.81,91) and (184,102.19) .. (184,116) .. controls (184,129.81) and (172.81,141) .. (159,141) .. controls (145.19,141) and (134,129.81) .. (134,116) -- cycle ;
\draw    (159,91) -- (159,141) ;

\draw    (273,41) -- (273,91) ;

\draw    (468,90) -- (468,140) ;

\draw    (184,116) -- (248,66) ;

\draw    (184,116) -- (252,162) ;

\draw   (346,64) .. controls (346,50.19) and (357.19,39) .. (371,39) .. controls (384.81,39) and (396,50.19) .. (396,64) .. controls (396,77.81) and (384.81,89) .. (371,89) .. controls (357.19,89) and (346,77.81) .. (346,64) -- cycle ;
\draw    (371,39) -- (371,89) ;

\draw    (298,66) -- (346,64) ;

\draw   (347,162) .. controls (347,148.19) and (358.19,137) .. (372,137) .. controls (385.81,137) and (397,148.19) .. (397,162) .. controls (397,175.81) and (385.81,187) .. (372,187) .. controls (358.19,187) and (347,175.81) .. (347,162) -- cycle ;
\draw    (372,137) -- (372,187) ;

\draw    (302,162) -- (347,162) ;

\draw    (396,64) -- (443,115) ;

\draw    (397,162) -- (443,115) ;

\draw (478,116) node  [align=left] {id};
\draw (172,115) node  [align=left] {id};
\draw (281,65) node [rotate=-270] [align=left] {{\small ReLU}};
\draw (286,160) node [rotate=-270] [align=left] {{\small ReLU}};
\draw (211,77) node  [align=left] {1};
\draw (210,148) node  [align=left] {\mbox{-}1};
\draw (381,65) node  [align=left] {id};
\draw (380,160) node [rotate=-268.41] [align=left] {$\displaystyle e^{a} -1$};
\draw (320,51) node  [align=left] {$\displaystyle 1/\alpha $};
\draw (326,147) node  [align=left] {$\displaystyle -1/\alpha $};
\draw (422,76) node  [align=left] {$\displaystyle \beta $};
\draw (426,145) node  [align=left] {$\displaystyle \beta $};

\end{tikzpicture}

%% file: FIGURE/ABU.tex
\tikzset{every picture/.style={line width=0.75pt}} 

\begin{tikzpicture}[x=0.75pt,y=0.75pt,yscale=-1,xscale=1]

\draw   (248,66) .. controls (248,52.19) and (259.19,41) .. (273,41) .. controls (286.81,41) and (298,52.19) .. (298,66) .. controls (298,79.81) and (286.81,91) .. (273,91) .. controls (259.19,91) and (248,79.81) .. (248,66) -- cycle ;
\draw   (251,128) .. controls (251,114.19) and (262.19,103) .. (276,103) .. controls (289.81,103) and (301,114.19) .. (301,128) .. controls (301,141.81) and (289.81,153) .. (276,153) .. controls (262.19,153) and (251,141.81) .. (251,128) -- cycle ;
\draw    (276,103) -- (276,153) ;

\draw   (368,129) .. controls (368,115.19) and (379.19,104) .. (393,104) .. controls (406.81,104) and (418,115.19) .. (418,129) .. controls (418,142.81) and (406.81,154) .. (393,154) .. controls (379.19,154) and (368,142.81) .. (368,129) -- cycle ;
\draw   (134,126) .. controls (134,112.19) and (145.19,101) .. (159,101) .. controls (172.81,101) and (184,112.19) .. (184,126) .. controls (184,139.81) and (172.81,151) .. (159,151) .. controls (145.19,151) and (134,139.81) .. (134,126) -- cycle ;
\draw    (159,101) -- (159,151) ;

\draw    (273,41) -- (273,91) ;

\draw    (393,104) -- (393,154) ;

\draw    (184,126) -- (248,66) ;

\draw    (184,126) -- (251,128) ;

\draw   (252,197) .. controls (252,183.19) and (263.19,172) .. (277,172) .. controls (290.81,172) and (302,183.19) .. (302,197) .. controls (302,210.81) and (290.81,222) .. (277,222) .. controls (263.19,222) and (252,210.81) .. (252,197) -- cycle ;
\draw    (277,172) -- (277,222) ;

\draw    (184,126) -- (252,197) ;

\draw    (298,66) -- (368,129) ;

\draw    (301,128) -- (368,129) ;

\draw    (302,197) -- (368,129) ;

\draw (403,130) node  [align=left] {id};
\draw (172,125) node  [align=left] {id};
\draw (281,65) node [rotate=-270] [align=left] {$\displaystyle f_{1}$};
\draw (285,126) node [rotate=-270] [align=left] {$\displaystyle f_{2}$};
\draw (218,84) node  [align=left] {1};
\draw (286,195) node [rotate=-270] [align=left] {$\displaystyle f_{k}$};
\draw (269,161) node [rotate=-270] [align=left] {$\displaystyle \dotsc $};
\draw (230,116) node  [align=left] {1};
\draw (226,156) node  [align=left] {1};
\draw (345,91) node  [align=left] {$\displaystyle \alpha _{1}$};
\draw (334,117) node  [align=left] {$\displaystyle \alpha _{2}$};
\draw (335,143) node  [align=left] {$\displaystyle \alpha _{k}$};

\end{tikzpicture}

%% file: FIGURE/APL.tex
\tikzset{every picture/.style={line width=0.75pt}} 

\begin{tikzpicture}[x=0.75pt,y=0.75pt,yscale=-1,xscale=1]

\draw   (248,27) .. controls (248,13.19) and (259.19,2) .. (273,2) .. controls (286.81,2) and (298,13.19) .. (298,27) .. controls (298,40.81) and (286.81,52) .. (273,52) .. controls (259.19,52) and (248,40.81) .. (248,27) -- cycle ;
\draw   (251,128) .. controls (251,114.19) and (262.19,103) .. (276,103) .. controls (289.81,103) and (301,114.19) .. (301,128) .. controls (301,141.81) and (289.81,153) .. (276,153) .. controls (262.19,153) and (251,141.81) .. (251,128) -- cycle ;
\draw    (276,103) -- (276,153) ;

\draw   (368,129) .. controls (368,115.19) and (379.19,104) .. (393,104) .. controls (406.81,104) and (418,115.19) .. (418,129) .. controls (418,142.81) and (406.81,154) .. (393,154) .. controls (379.19,154) and (368,142.81) .. (368,129) -- cycle ;
\draw   (134,126) .. controls (134,112.19) and (145.19,101) .. (159,101) .. controls (172.81,101) and (184,112.19) .. (184,126) .. controls (184,139.81) and (172.81,151) .. (159,151) .. controls (145.19,151) and (134,139.81) .. (134,126) -- cycle ;
\draw    (159,101) -- (159,151) ;

\draw    (273,2) -- (273,52) ;

\draw    (393,104) -- (393,154) ;

\draw    (184,126) -- (248,27) ;

\draw    (184,126) -- (251,128) ;

\draw   (252,197) .. controls (252,183.19) and (263.19,172) .. (277,172) .. controls (290.81,172) and (302,183.19) .. (302,197) .. controls (302,210.81) and (290.81,222) .. (277,222) .. controls (263.19,222) and (252,210.81) .. (252,197) -- cycle ;
\draw    (277,172) -- (277,222) ;

\draw    (184,126) -- (252,197) ;

\draw    (298,27) -- (368,129) ;

\draw    (301,128) -- (368,129) ;

\draw    (302,197) -- (368,129) ;

\draw   (251,271) .. controls (251,257.19) and (262.19,246) .. (276,246) .. controls (289.81,246) and (301,257.19) .. (301,271) .. controls (301,284.81) and (289.81,296) .. (276,296) .. controls (262.19,296) and (251,284.81) .. (251,271) -- cycle ;
\draw    (276,246) -- (276,296) ;

\draw    (184,126) -- (251,271) ;

\draw    (368,129) -- (301,271) ;

\draw (403,130) node  [align=left] {id};
\draw (172,125) node  [align=left] {id};
\draw (281,26) node [rotate=-270] [align=left] {ReLU};
\draw (285,126) node [rotate=-270] [align=left] {ReLU};
\draw (211,64) node  [align=left] {1};
\draw (286,195) node [rotate=-270] [align=left] {ReLU};
\draw (272,235) node [rotate=-270] [align=left] {$\displaystyle \dotsc $};
\draw (230,116) node  [align=left] {\mbox{-}1};
\draw (226,156) node  [align=left] {\mbox{-}1};
\draw (334,117) node  [align=left] {$\displaystyle w_{u,1}$};
\draw (285,269) node [rotate=-270] [align=left] {ReLU};
\draw (338,61) node  [align=left] {1};
\draw (262,129) node [scale=1.0,rotate=-270] [align=left] {$\displaystyle b_{u,1}$};
\draw (264,199) node [scale=1.0,rotate=-270] [align=left] {$\displaystyle b_{u,2}$};
\draw (262,271) node [scale=1.0,rotate=-270] [align=left] {$\displaystyle b_{u,k}$};
\draw (326,150) node  [align=left] {$\displaystyle w_{u,2}$};
\draw (351,204) node  [align=left] {$\displaystyle w_{u,k}$};
\draw (230,206) node  [align=left] {\mbox{-}1};

\end{tikzpicture}

%% file: TAB/setup.tex
\begin{table}[h!]
\centering
\resizebox{\textwidth}{!}{%
\begin{tabular}{c|c|c}
\hline
\hline
\textbf{Reference paper} & \textbf{Setup architecture(s)} & \textbf{Dataset(s)} \\ \hline
Maxout \citep{goodfellow2013} & owns & \begin{tabular}[c]{@{}l@{}}MNIST \\ CIFAR 10/100\end{tabular} \\ \hline
NIN \citep{lin2013} & owns & \begin{tabular}[c]{@{}l@{}}SVHN \\ MNIST \\ CIFAR 10/100\end{tabular} \\ \hline
ProbMaxout \citep{springenberg2013} & based on \citep{goodfellow2013} & \begin{tabular}[c]{@{}l@{}}SVHN \\ CIFAR 10/100\end{tabular} \\ \hline
BN-MIN \citep{chang2015} & based on \citep{lin2013} & \begin{tabular}[c]{@{}l@{}}SVHN \\ MNIST \\ CIFAR 10/100\end{tabular} \\ \hline
PRELU \citep{he2015} & based on VGG-19 \citep{simonyan2014} & ImageNet \\ \hline
APL \citep{agostinelli2014} & based on \citep{srivastava2014} & \begin{tabular}[c]{@{}l@{}}CIFAR 10/100 \\ Higgs to $\tau^+-\tau^-$ \citep{baldi2015}\end{tabular} \\ \hline
S-Shaped \citep{jin2016} & \begin{tabular}[c]{@{}l@{}}based on \citep{lin2013} and \\ GoogleNet \citep{szegedy2015}\end{tabular} & \begin{tabular}[c]{@{}l@{}}MNIST \\ CIFAR 10/100 \\ ImageNet\end{tabular} \\ \hline
H.\& K. \citep{harmon2017} & \begin{tabular}[c]{@{}l@{}}owns and \\ Lasagne ResNet (http://github.com/Lasagne)\end{tabular} & \begin{tabular}[c]{@{}l@{}}MNIST\\ ISOLET \\ CIFAR 10/100\end{tabular} \\ \hline
\citep{eisenach2016} & based on \citep{srivastava2014} & \begin{tabular}[c]{@{}l@{}}MNIST \\ CIFAR10\end{tabular} \\ \hline
Swish \citep{ramachandran2017} & \begin{tabular}[c]{@{}l@{}}based on ResNets \citep{he2016,zagoruyko2016},\\ DenseNet \citep{huang2016},\\ Inception \citep{szegedy2017},\\ MobileNets \citep{howard2017,zoph2018} and\\ Transformer \citep{vaswani2017}\end{tabular} & \begin{tabular}[c]{@{}l@{}}CIFAR 10/100 \\ ImageNet, \\ WMT2014\end{tabular} \\ \hline
GA/MA/HA \citep{qian2018} & based on \citep{lin2013} & \begin{tabular}[c]{@{}l@{}}MNIST \\ CIFAR 10/100 \\ ImageNet\end{tabular} \\ \hline
ABU \citep{sutfeld2018} & owns & CIFAR 10/100 \\ \hline
FReLU \citep{qiu2018} & \begin{tabular}[c]{@{}l@{}}owns and\\ based on ResNets (https://github.com/facebook)\end{tabular} & CIFAR 10/100 \\ \hline
MMN \citep{sun2018} & \begin{tabular}[c]{@{}l@{}}owns and\\ based on \citep{lin2013}\end{tabular} & \begin{tabular}[c]{@{}l@{}}CIFAR 10/100 \\ ImageNet\end{tabular} \\ \hline
LUTU \citep{wang2018} & based on ResNets (https://github.com/facebook) & \begin{tabular}[c]{@{}l@{}}CIFAR10 \\ ImageNet\end{tabular} \\ \hline
PELU \citep{trottier2017} & \begin{tabular}[c]{@{}l@{}}based on ResNets \citep{shah2016},\\ \citep{lin2013},\\ AllCNN \citep{springenberg2015},\\ Overfeat \citep{sermanet2014}\end{tabular} & \begin{tabular}[c]{@{}l@{}}CIFAR 10/100\\ ImageNet\end{tabular} \\ \hline
KAF \citep{scardapane2018} & \begin{tabular}[c]{@{}l@{}}owns,\\ based on \citep{baldi2014} and\\ VGG \citep{simonyan2014}\end{tabular} & \begin{tabular}[c]{@{}l@{}}Sensorless\citep{dua2017}\\ SUSY \citep{baldi2014} \\ MNIST\\ Fashion MNIST \\ CIFAR 10/100\end{tabular} \\ \hline
VAF \citep{apicella2019} & \begin{tabular}[c]{@{}l@{}}owns,\\ based on \citep{lin2013} and\\ \citep{scardapane2018}\end{tabular} & \begin{tabular}[c]{@{}l@{}}Sensorless \citep{dua2017},\\ MNIST\\ Fashion MNIST\\ CIFAR10\\ others from UCI \citep{dua2017}\end{tabular} \\ \hline
LWTA \citep{srivastava2013compete} & \begin{tabular}[c]{@{}l@{}}owns\end{tabular} & \begin{tabular}[c]{@{}l@{}} MNIST\\Amazon Sentiment Analysis \citep{blitzer2007biographies}\end{tabular} \\ 
\hline
\hline
\end{tabular}%
}
\caption{The setup architectures used by some works and the dataset used for the experiments made.}
\label{tab:setup}
\end{table}

%% file: conclusions.tex
\section{Conclusions}
\label{sec:conclusions}
In this paper, we have discussed the most common activation functions \textcolor{black}{presented in the literature,} focusing on the trainable ones and proposing a possible taxonomy to distinguish them. We have divided the proposed functions into two main categories: fixed shape and trainable shape. 
In the second class of activation functions \textcolor{black}{two different families of trainable activation functions have been individuated:} \textit{parameterized standard functions} and \textit{functions based on ensemble methods}. 
The latter has been refined further by isolating another activation function family: \textit{linear combination of one-to-one functions}. 
This taxonomy shows the great variety of activation functions proposed \textcolor{black}{in literature,} and several works on trainable activation functions report substantial performance improvements in comparison with equivalent neural network architectures equipped with classical fixed-shape activation functions as ReLU or sigmoid.
However, it must be kept in mind that the activation functions are not the only entities that determine the performances of a neural network. 
Other hyperparameters, such as the number of neurons and the way they are arranged between them or the weights initialization protocol, can be decisive for the performance of the network, together with the hyper-parameter values required by the learning algorithms. 
Furthermore, even if the experiments are conducted on the same datasets, these may have been pre-processed in different ways (e.g., ZCA \citep{bell1997} or data-augmentation \citep{perez2017}) which can in turn condition the results. So, it can be challenging to make a comparison, in terms of network performances, among the activation \textcolor{black}{functions that have been proposed in literature,} since the experiments are often conducted using different experimental setups. Table \ref{tab:accfun} shows the performances in terms of accuracy reported in several works on the most commonly used datasets.
As an indication, we report just the best values obtained, without taking care of the different architectures or experimental setup used \textcolor{black}{in the different works.}

By a more in-depth analysis, it is important to note that several proposed trainable activation functions can be expressed in terms of linear combination of fixed non-linear functions, so they can be expressed in turn as subnetworks nested in the main architectures, as explicitly reported in the models proposed in \citep{qian2018,apicella2019}. 
In Figures \ref{fig:swishnn} \ref{fig:srelunn} and \ref{fig:nnform}, we try to draw possible equivalent implementations of some proposed functions in terms of neural (sub)networks architectures.
Looking at these functions in this respect, it is easy to see that every neural network architecture equipped with one of these trainable functions can be modelled with an equivalent deeper architecture equipped using just fixed-shape functions. This could lead to the observation that several architectures that use trainable activation functions could reach similar results using deeper architectures possibly imposing some constraints on specific weight layers such as weight sharing without needing trainable activation functions. 
Thus, neural networks with trainable activation functions can be cheaper in terms of computational complexity since there are fewer parameters to control (as weights with fixed values), on the other hand, the performances in accuracy terms are potentially comparable to deeper architectures without trainable activation functions.
It is worth to point out that some authors, see for example \citep{apicella2019,scardapane2018}, suggest defining trainable activation functions able to satisfy several desirable properties which include a high expressive power of the trainable activation functions, no external parameter or learning process in addition to the classical ones for neural networks, and the possibility to use classical regularisation methods. 
These properties can be easily satisfied when trainable activation functions are explicitly expressed in terms of subnetworks. 
\textcolor{red}{
Notice that we are not proposing to substitute, in general, trainable activation functions with their corresponding equivalent representations in terms of sub-networks. On the other hand, we
think that these equivalent representations might better emphasize similarities and differences among the various trainable activation functions insofar as they are represented using the same formalism, and  all the functional properties remain unchanged during the learning stage as they share both the same parameters and
the same input-output functional relationship. It is worth pointing out that, however, the equivalent representations of trainable activation functions as sub-networks could be more expensive in terms of computational costs.
}

%% file: TAB/results.tex
\begin{table}[t]
\resizebox{\textwidth}{!}{%
\begin{tabular}{cc|cccc}
\hline
\hline
\textbf{} & \textbf{} &  &  & \textbf{Accuracy \%} & \textbf{} \\ 
\textbf{} & \textbf{} & \textbf{SVHN} & \textbf{MNIST} & \textbf{CIFAR10} & \textbf{CIFAR100} \\ \hline
\textbf{Fixed shape} & \textit{sigmoid} &  & 97.9 \cite{pedamonti2018}; &  &  \\ \hline
\textit{classic} & \textit{tanh} &  & 98.21 \cite{eisenach2016} &  &  \\ \hline
 & \textit{softplus} &  &  & 94.9 \cite{ramachandran2017} & 83.7 \cite{ramachandran2017} \\ \hline
\textit{Rectifier-based} & ReLU &  & \begin{tabular}[c]{@{}l@{}}98.0 \cite{pedamonti2018};\\ 99.53 \cite{jin2016};\\ 99.16 \cite{eisenach2016};\\ 99.1 \cite{apicella2019} \\ 99.15 \cite{scardapane2018} \end{tabular} & \begin{tabular}[c]{@{}l@{}}87.55 \cite{xu2015};\\ 92.27 \cite{jin2016};\\ 94.59 \cite{trottier2017};\\ 91.51 \cite{qian2018};\\ 84.8 \cite{eisenach2016};\\ 95.3 \cite{ramachandran2017};\\ 85.7\cite{apicella2019}\end{tabular} & \begin{tabular}[c]{@{}l@{}}57.1 \cite{xu2015};\\ 67.25 \cite{jin2016};\\ 75.45 \cite{trottier2017};\\ 64.42 \cite{qian2018}\\ 83.7 \cite{ramachandran2017}\end{tabular} \\ \hline
\textbf{} & LReLU &  & \begin{tabular}[c]{@{}l@{}}98.2 \cite{pedamonti2018};\\ 99.58 \cite{jin2016};\end{tabular} & \begin{tabular}[c]{@{}l@{}}88.8 \cite{xu2015};\\ 92.3 \cite{jin2016};\\ 92.32 \cite{qian2018};\\ 95.6 \cite{ramachandran2017}\end{tabular} & \begin{tabular}[c]{@{}l@{}}59.58 \cite{xu2015};\\ 67.3 \cite{jin2016}:\\ 64.72 \cite{qian2018};\\ 83.3 \cite{ramachandran2017}\end{tabular} \\ \hline
\textbf{} & RReLU &  &  & 88.81 \cite{xu2015} & 59.75 \cite{xu2015} \\ \hline
\textbf{} & ELU &  & 98.3 \cite{pedamonti2018}; & \begin{tabular}[c]{@{}l@{}}93.45 \cite{clevert2015};\\ 94.01 \cite{trottier2017};\\ 92.16 \cite{qian2018};\\ 94.4 \cite{ramachandran2017}\end{tabular} & \begin{tabular}[c]{@{}l@{}}75.72 \cite{clevert2015};\\ 74.92 \cite{trottier2017};\\ 64.06 \cite{qian2018};\\ 80.6 \cite{ramachandran2017}\end{tabular} \\ \hline
 & \textit{AGSig (Swish-1)} &  &  & 95.5 \cite{ramachandran2017} & 83.8 \cite{ramachandran2017} \\ \hline
 & \textit{} &  &  &  &  \\ \hline
\textbf{Learnable shape} &  &  &  &  &  \\ \hline
\textit{Quasi fixed} & PReLU &  & 99.59 \cite{jin2016} & \begin{tabular}[c]{@{}l@{}}88.21 \cite{xu2015};\\ 92.32 \cite{jin2016};\\ 94.64 \cite{trottier2017};\\ 95.1 \cite{ramachandran2017}\end{tabular} & \begin{tabular}[c]{@{}l@{}}58.37 \cite{xu2015};\\ 67.33 \cite{jin2016};\\ 74.5 \cite{trottier2017};\\ 81.5 \cite{ramachandran2017}\end{tabular} \\ \hline
\textbf{} & PELU &  &  & 94.64 \cite{trottier2017} & 75.45 \cite{trottier2017} \\ \hline
\textbf{} &  &  &  &  &  \\ \hline
\textbf{} & \textit{} &  &  &  &  \\ \hline
\textit{Interpolated} & LuTU &  &  & 94.22 \cite{wang2018} &  \\ \hline
\textbf{} &  &  &  &  &  \\ \hline
\textit{Ensembled} & Gated Act. &  &  & 92.65 \cite{qian2018} & 65.75 \cite{qian2018} \\ \hline
\textbf{} & Mixed Act. &  &  & 92.6 \cite{qian2018} & 65.44 \cite{qian2018} \\ \hline
\textbf{} & hierarc. Act. &  &  & 92,99 \cite{qian2018} & 66.23 \cite{qian2018} \\ \hline
\textbf{} & \textit{Harmon Klabjan} &  & 99.40 \cite{harmon2017} &  & 74.20 \cite{harmon2017} \\ \hline
\textbf{} & SReLU &  & 99.65 \cite{jin2016} & 93.02 \cite{jin2016} & 70.09 \cite{jin2016} \\ \hline
\textbf{} & \textit{APL} &  & 99.31 \cite{scardapane2018} & 92.49 \cite{agostinelli2014} & 69.17 \cite{agostinelli2014} \\ \hline
\textbf{} & NPF &  & 99.31 \cite{eisenach2016} & 86.44 \cite{eisenach2016} &  \\ \hline
\textbf{} & Swish &  &  & 95.5 \cite{ramachandran2017} & 83.9 \cite{ramachandran2017} \\ \hline
\textbf{} & VAF &  & 99.5  \cite{apicella2019} &  81.2 \cite{apicella2019} & \\ \hline
\textbf{} & KAF &  & \begin{tabular}[c]{@{}l@{}}99.43  \cite{scardapane2018}\\99.5 \cite{apicella2019}\\ \end{tabular}& \begin{tabular}[c]{@{}l@{}} ~84.0 \cite{scardapane2018}\\80.2 \cite{apicella2019}\end{tabular} & ~52.0 \cite{scardapane2018}\\ \hline
\textbf{} &  &  &  &  &  \\ \hline
\textbf{} &  &  &  &  &  \\ \hline
\textbf{Changing T.F.} & \textit{Maxout} & 97.53 \cite{goodfellow2013} & 99.55 \cite{goodfellow2013} & 90.62 \cite{goodfellow2013} & 61.43 \cite{goodfellow2013} \\ \hline
\textbf{} & \textit{Probout} & 97.61 \cite{springenberg2013} &  & 88.65 \cite{springenberg2013} & 61.86 \cite{springenberg2013} \\ \hline
\textbf{} & MMN &  &  & 92,34 \cite{sun2018} & 66.76 \cite{sun2018} \\ \hline
\textbf{} & NIN & 97.65 \cite{lin2013} & \begin{tabular}[c]{@{}l@{}}99.53 \cite{lin2013}\\99.6 \cite{apicella2019}\end{tabular}& \begin{tabular}[c]{@{}l@{}}91.19 \cite{lin2013}\\76.3 \cite{apicella2019}\end{tabular} & 64.32 \cite{lin2013} \\ \hline
\textbf{} & CIC &  &  & 91.54 \cite{pang2016} & 68.6 \cite{pang2016} \\ \hline
\textbf{} & MIN & 98.19 \cite{chang2015} & 99.76 \cite{chang2015} & 92.15 \cite{chang2015} & 71.14 \cite{chang2015} \\ \hline

\textbf{} & LSWA & & 99.43 \cite{srivastava2013compete} & & \\ \hline
\textbf{} & \textit{} &  &  &  &  \\ 
\hline
\hline
\end{tabular}%
}
\caption{Accuracy on some datasets of the most common activation functions in literature; here, we report the best values reported without taking into account the differences between the architectures used.}
\label{tab:accfun}
\end{table}